%% file: article_arxiv.tex
\documentclass[journal]{IEEEtran}
\usepackage[T1]{fontenc}
\usepackage{lmodern}
\usepackage{amsmath,amssymb,booktabs,graphicx}
\usepackage[hidelinks]{hyperref}

\usepackage{float,needspace,placeins,subcaption}
\usepackage{algorithm,algpseudocode}
\usepackage{tabularx} 
\usepackage{tikz}
\usepackage{soul}
\usetikzlibrary{arrows.meta,positioning,shapes.geometric}
\usepackage[hidelinks]{hyperref}
\hypersetup{
	pdftitle={A State Based Dispatch Controller for Hospital Delivery Robots with Shared Human and Infrastructure Resources},
	pdfauthor={Krzysztof Siwek and Aleksandra Świetlicka},
	pdfsubject={Formal dispatch control and discrete event simulation for hospital delivery robots},
	pdfkeywords={hospital service robots, fleet management, resource aware dispatch, human robot systems, discrete event simulation, decision support}
}
\graphicspath{{images/}}
\title{A State Based Dispatch Controller for Hospital Delivery Robots with Shared Human and Infrastructure Resources}
\author{Krzysztof Siwek\\Faculty of Electrical Engineering\\Warsaw University of Technology\\Koszykowa 75, 00-662 Warszawa\\email:krzysztof.siwek@pw.edu.pl\\
	\vspace{1em}
	Aleksandra \ifmmode\acute{S}\else\'{S}\fi{}wietlicka\\Faculty of Control, Robotics and Electrical Engineering\\Pozna\'n University of Technology\\Piotrowo 3a, 61-138 Poznan\\email:aleksandra.swietlicka@put.poznan.pl}
\date{}

\begin{document}
\maketitle
\begin{abstract}
Hospital delivery robots share staff, elevators, charging stations, and cleaning facilities with clinical operations. We formulate adaptive resource aware dispatch (ARAD), a deterministic state feedback controller that maps each arriving request and the observable system state to human or robot assignment. Its admissibility conditions represent human availability, deadline slack, elevator pressure, human queue pressure, service station pressure, and a fleet reserve. We evaluate ARAD in a discrete event simulation based on 33,079 transport requests from the University Medical Center Freiburg and a synthetic task mix derived from MIMIC demo event counts. The model schedules robot loading, receiving, coordination, cleaning, and recovery on individual transporter shifts. It also enforces battery feasibility and shared elevator capacity. In the earlier experiment, a recalibrated human roster was compared with one, two, and four robot fleets in 30 paired replications. Two ARAD robots increased overall late or incomplete requests by 0.37 percentage points (95 percent Monte Carlo confidence interval, 0.26 to 0.49) and urgent lateness by 0.82 points (0.70 to 0.95). Human idle time in intervals of at least 10 minutes decreased by 11.5 minutes per 100 nominal beds per day, while robot support consumed 56.4 minutes on the same scale. A deadline feasible benchmark did not improve on the human roster. The formal controller and coupled resource model make the assignment logic reproducible, while the results show that an admissible robot mission need not improve system performance when human support and building congestion are included.
 The updated operational study adds location aware travel and assignment, two full scenario grids totaling 2,160 runs, and 120 further deadline benchmark runs. In one high roster scenario, increasing assumed elevator capacity from two to four servers lowers human only lateness from 55.11 percent to 3.18 percent. The dispatch rules do not show a stable service gain across the updated grids. These conditional comparisons remain sensitive to unobserved staffing, deadlines, and building operations.
\end{abstract}
\noindent\textbf{Keywords:} hospital service robots; fleet management; resource aware dispatch; human robot systems; discrete event simulation; decision support.

\section{Introduction}
Hospital delivery robots share corridors and elevators with people, making their operation part of a broader human--robot interaction setting in which robotic systems must account for both human behavior and the shared environment \cite{swietlicka2025survey} Mobile robots can perform some of these journeys, but service depends on more than travel speed. Robots also need elevators, charging stations, and cleaning facilities. A review and an observational study of robotics in nursing describe applications and implementation challenges \cite{babalola2024,sommer2024}. For transport planning, another question is whether moving work to robots changes the timing of staff availability in a useful way.

Simulation, scheduling, and fleet control for hospital delivery robots are established research areas. Bhosekar and colleagues modeled hospital inventory and material handling \cite{bhosekar2023}. Cheng and colleagues studied stochastic scheduling of autonomous mobile robots with time windows and hospital costs \cite{cheng2023}. Guzm\'an Ortiz and colleagues described a fleet management architecture for healthcare environments \cite{guzman2021}. Ge and colleagues used discrete event simulation to study automated guided vehicle systems in operating theaters \cite{ge2025}. Recent hospital pilots have also evaluated a modular logistics robot in clinical environments \cite{dei2026}. These studies show that the main research gap is not whether hospital robots can be simulated or scheduled. The unresolved question is how dispatch choices, shared building resources, and human support work jointly affect service and staff availability.

This study contributes a formal state and decision model for adaptive resource aware dispatch, a simulation architecture that couples robot missions to human support and shared infrastructure, and a reproducible evaluation against a deadline feasible controller and a feasible human roster. ARAD is a transparent rule based controller rather than a learned or globally optimized policy. Its purpose is to make every robot admission decision traceable to observable state and explicit constraints. The experiments report deadline performance, transporter idle interval structure, modeled energy, support burden, charging, cleaning, elevator traffic, and movement between assignments. The central question is conditional: when does a robot assignment remain beneficial after the resources required to complete that mission are represented explicitly?

This extended article retains the earlier experiments and adds the updated operational study reported in the current IEEE manuscript. The earlier primary model omits travel to pickups and uses a roster screened under that assumption. Its repositioning experiment is a separate stress case. The updated study instead includes worker and robot locations, travel to pickups and support handoffs, and nearest available assignment in every main scenario. The two studies have different operating models and rosters. Their absolute outcomes should not be treated as estimates from one unchanged experiment. Sections explicitly identify the earlier experiments and the updated operational analysis.

\input{literature_review.tex}
\input{robotics_context.tex}

\section{Methods for Earlier Experiments}
\subsection{Study design, scope, and empirical inputs}
The Python and SimPy model uses a seven day warmup, 30 measured days, and a further day to complete outstanding requests. Service outcomes include only requests created during measurement, and idle intervals are clipped to that window. A run length analysis compares zero, seven, and 14 warmup days and 30 and 60 measured days. The study is a conditional scenario analysis. It estimates differences between dispatch rules within the implemented model and does not estimate a population effect for hospitals. Reporting is organized around the STRESS framework \cite{monks2019}.

The Freiburg dataset contains 33,079 requests across 92 recorded dates, a network of 6,707 nodes and 13,834 edges, and 1,913 observed origin and destination pairs \cite{hess2026}. Each simulated day samples a recorded daily request profile with replacement. Replay preserves origins, destinations, and within-day request times to minute resolution. It does not preserve calendar order, weekday patterns, seasonal variation, or dependence between successive days. The dataset provides requests and routes, but it does not provide completion times, actual staffing, transported contents, priorities, or deadlines.

Clinical contents and priorities are not observed. Synthetic classes are assigned independently of route and time using relative event counts from MIMIC-IV demo and MIMIC-IV-ED demo, both version 2.2 \cite{mimicdemo,mimiced,johnson2023,physionet2026}. Medication weights combine 1,082 dispensing records and 18,087 prescriptions. Specimen weights use 107,727 laboratory records, supply weights use 1,190 transfers, and instrument weights use the integer part of one quarter of 275 admissions, or 68. Normalization gives Table~\ref{tab:task-assumptions}. These source events are not comparable transport units. A laboratory row need not represent a distinct specimen journey, and a prescription need not produce a separate delivery. The resulting task mix is a synthetic scenario, not a calibration to observed hospital transport. MIMIC arrival weights and acuity probabilities are not used by the Freiburg daily replay; assigned task class alone determines urgency.

\begin{table*}[htb]
\centering\small
\caption{Synthetic task assumptions. Priority 1 defines urgent requests. Handling times are fixed inputs, not transport observations.}
\label{tab:task-assumptions}
\begin{tabular}{lrrrrr}
\toprule
Class & Share (\%) & Priority & Deadline (min) & Loading (min) & Receiving (min) \\
\midrule
Medication & 14.958 & 2 & 35 & 1.4 & 1.0 \\
Specimen & 84.061 & 1 & 25 & 1.0 & 0.8 \\
Supply & 0.929 & 4 & 120 & 2.0 & 1.5 \\
Instrument & 0.053 & 2 & 45 & 2.5 & 1.5 \\
\bottomrule
\end{tabular}
\end{table*}

The virtual hospital has 200 nominal beds for normalization. This is not an estimate of the Freiburg campus bed count, and demand is not scaled to an observed census. Values per 100 beds use a fixed model denominator rather than an empirical relationship between beds and workload.

\subsection{Resources and model boundaries}
Transporters work individual eight hour shifts. The revised staffing experiment screened night, day, and evening counts, then confirmed schedules near the minimum feasible staffed hours in 30 replications. The primary schedule has two night, eight day, and seven evening transporters, totaling 136 staffed hours per day. It meets assumed criteria of no more than 10 percent overall late or incomplete requests, no more than 5 percent urgent late or incomplete requests, fewer than 0.5 incomplete requests on average, and human utilization no greater than 0.85. This roster does not reconstruct actual Freiburg staffing. Human requests and robot support assignments share one priority queue ordered by priority, deadline, and insertion sequence. Service is not preempted, and workers finish an assigned activity even beyond shift end.

Journey duration combines fixed handling times, shortest route distance, mode specific speed, elevator travel, and stochastic congestion. Human and robot speeds are 72 and 54 meters per minute. Explicit elevator contention aggregates all elevator service on a route into one request to a shared bank of two servers. Individual cabins and separate floor queues are not represented. The primary model has no passenger stream, while sensitivity scenarios add independent Poisson arrivals at six or 12 passenger requests per hour. The primary model does not add travel from the previous destination to the next pickup. A separate stress experiment tracks worker and robot locations and adds this travel through the empirical route graph. Travel to charging and cleaning stations, obstacle avoidance, localization errors, and pedestrian interactions remain omitted.

In the primary experiment, each robot has a 4 kWh battery and starts charging after a mission if charge is at or below 25 percent. Two 2 kW chargers restore full charge with efficiency 0.90. Planned mission time consumes 0.025 kWh per minute. Before a mission begins, the model checks whether the current charge is sufficient and charges the robot when needed. A mission that exceeds full battery capacity is rejected as infeasible. Energy excludes elevator waiting, cleaning, standby, recovery, and travel to support stations; reported kWh are modeled mission energy rather than grid electricity.

Specimen and instrument missions require seven minutes of cleaning at two shared stations. This models downtime, not contamination or infection risk. Cleaning also requires two minutes of transporter support. Each mission avoids modeled failure with probability 0.98. Failure adds exponential recovery with mean 12 minutes. Coordination is $2\exp(Z)$ minutes, where $Z\sim N(0,0.3^2)$, giving a mean of about 2.09 minutes before recovery. Loading, receiving, coordination, recovery, and cleaning support are inserted into the same human queue as transport requests. A robot remains occupied while waiting for loading, receiving, coordination, or recovery support. These assumptions make support timing explicit, although they are not calibrated observations.

\input{system_architecture.tex}

\subsection{Dispatch rules}
Overflow assigns a request to an available robot when no human is idle or at least one request is waiting for humans. Priority dispatch assigns nonurgent requests to an available robot and urgent requests if estimated robot duration is no more than 75 percent of deadline slack. With batteries active, robots are taken from an idle store. This differs from the initial model without batteries, where priority requests could wait for a robot.

Adaptive resource aware dispatch is a deterministic controller that responds to the observed system state. It is not learned and does not solve a horizon wide optimization problem. Urgent requests may be assigned to a robot when the duration estimate, inflated by elevator pressure, fits within remaining slack. Nonurgent assignments also require sufficient human queue pressure, acceptable charging and cleaning pressure, and a fleet reserve. The deadline feasible benchmark retains robot availability, human availability, and deadline feasibility but omits the reserve and nonurgent pressure gates. Separate ablations remove one ARAD gate at a time. Section~\ref{sec:formal-controller} specifies the state, inputs, objectives, admissibility conditions, and assignment algorithm.

\input{dispatch_formalization.tex}

\subsection{Outcomes and statistical analysis}
The primary outcome is the fraction of measurement requests completed after their deadline or still incomplete after clearance. This choice was not prospectively registered. Urgent lateness uses the same definition for priority 1. Let $\mathcal{I}_j$ contain worker $j$'s idle intervals after removing human delivery service intervals. The gross idle interval measure is
\begin{equation}
I_\tau=\frac{100}{BD}\sum_j\sum_{\ell\in\mathcal{I}_j}\ell\,\mathbf{1}(\ell\geq\tau),
\end{equation}
where $B=200$ nominal beds, $D=30$ measured days, and $\tau=10$ minutes. The entire qualifying interval is counted. Robot support is absent from $\mathcal{I}_j$, so this measures gross delivery idle time. The 10 minute threshold is an analytical convention rather than a clinically validated duration. Subtracting aggregate support minutes cannot reconstruct the fragmentation caused by support tasks.

The revised primary experiment uses one human only baseline and ARAD and deadline feasible policies with fleets of one, two, and four robots. Each scenario has 30 replications. Seeds 20261201 through 20261230 identify paired replications. Ablations, task mapping, and passenger elevator scenarios use 20 replications. Run length scenarios use 15, and repositioning stress scenarios use 10. A separate demand stream preserves paired demand. Operational draws share a master seed but their consumption order changes with policy, so they are not synchronized request by request.

For $n$ paired differences $d_r$, intervals are $\bar d\pm t_{0.975,n-1}s_d/\sqrt{n}$. Direct paired policy contrasts are also calculated. These pointwise Monte Carlo intervals condition on the empirical day pool and fixed inputs. They exclude uncertainty in task mapping, hospital sampling, model structure, and implementation. No multiplicity correction is applied; exploratory comparisons are descriptive. Earlier component experiments cannot be added together or treated as validation of the final ARAD policy.

\input{model_details.tex}

\section{Results from Earlier Experiments}
\input{revision_results.tex}

\section{Updated Operational Study: Model and Comparison Design}
\label{sec:updated-operational-study}
The updated study addresses the pickup travel limitation of the earlier primary model. It uses the full directed Freiburg pathway network for movement between assignments and schedules human support at its actual handoff location. The earlier experiments remain useful as development and sensitivity evidence, while the following design governs all results in the updated operational study. The proximity benchmark introduced here also differs from the earlier repositioning stress rule. No observed staffing, task contents, deadlines, support durations, or completion times became available between the two studies.

\subsection{What the request record measures}
The Freiburg release contains 33,079 requests on 92 recorded dates, 1,913 observed origin and destination pairs, and a network with 6,707 nodes and 13,834 directed edges \cite{hess2026}. Each simulated day samples a recorded daily profile with replacement. Within that day, the model preserves request times to the minute and the recorded endpoints. It does not preserve the sequence of calendar days or the relationship between one day and the next. Delivery routes use the shortest paths calculated from the released pathway graph. Travel between a worker's previous destination and a new pickup also uses that graph. Direct observed origin and destination estimates remain the reference for each delivery.

The requests concern small and medium goods, but the public file has no contents for individual orders. We assign classes independently of time and route. One scenario normalizes event counts from the MIMIC demonstration releases: specimens 84.06\%, medication 14.96\%, supplies 0.93\%, and instruments 0.05\% \cite{mimicdemo,mimiced}. Laboratory events and prescriptions are not transport trips. We call this the \emph{source count mix}, not a measured Freiburg task mix. A second, balanced scenario uses 40\% specimens, 30\% medication, 20\% supplies, and 10\% instruments. Neither scenario is a clinical estimate.

Table~\ref{updated-tab:tasks} gives the assumed handling times, priorities, and deadlines. Priority 1 is urgent. We vary the specimen deadline across 25, 45, and 60 minutes because specimens dominate the source count mix. The other deadlines remain fixed. The values are engineering probes, not observed standards of care.
\begin{table}[t]
\caption{Assumed task attributes. Times are minutes.}
\label{updated-tab:tasks}
\centering\footnotesize
\begin{tabular}{lrrrr}
\toprule
Class & Priority & Deadline & Load & Receive \\
\midrule
Specimen & 1 & 25, 45, 60 & 1.0 & 0.8 \\
Medication & 2 & 35 & 1.4 & 1.0 \\
Supply & 4 & 120 & 2.0 & 1.5 \\
Instrument & 2 & 45 & 2.5 & 1.5 \\
\bottomrule
\end{tabular}
\end{table}

\subsection{Operating scenarios}
The model has eight hour night, day, and evening shifts. A moderate roster contains 8, 30, and 24 transporters on those shifts. A high roster contains 15, 60, and 50. These are broad sensitivity cases, not estimates of employment at Freiburg. Every worker starts a shift at network node 2624, an arbitrary reference station. That rule is held fixed in every comparison. A worker finishes an assigned activity after shift end when necessary. The model uses 200 nominal beds only to normalize reported staff minutes; the value is not the campus census.

Humans move at an assumed 72 m/min and robots at 54 m/min. Distance travel is multiplied by a lognormal congestion factor with standard deviation parameter 0.18 and mean one before truncation at 0.65. Route elevator service is requested from a shared bank. The initial grid assumes two servers; a second full grid assumes four, and a human only sensitivity case assumes eight. These capacities describe one pooled bank for the entire network. They are not measurements of the hospital's elevators. Each ride contributes 2.2 minutes plus 0.25 minute per floor crossed. A route aggregates its elevator use into one bank request. Passenger arrivals, individual cabin direction, and corridor congestion from pedestrians are absent. These simplifications make the comparison reproducible but limit physical interpretation.

Two robots are available in each hybrid scenario. Each has a 4 kWh battery, a 25\% charge trigger, and access to two 2 kW chargers with assumed efficiency 0.90. Mission energy is 0.025 kWh per planned active minute. Specimen and instrument missions require seven minutes at one of two cleaning stations and two minutes of transporter support. A mission avoids a modeled failure with probability 0.98; a failure adds exponentially distributed recovery with a mean of 12 minutes. Coordination has a lognormal duration with scale two minutes. These are scenario values, not telemetry or infection control validation.

\subsection{Updated coupled dispatch model}
\subsection{Observable state and assignment estimates}
\label{updated-sec:formal-controller}
Request $i$ arrives at $a_i$ with origin $o_i$, destination $d_i$, priority $p_i$, absolute due time $\delta_i$, loading duration $L_i$, and receiving duration $R_i$. Priority 1 denotes urgent work. The state at an arrival contains the active and busy transporter counts $H_a$ and $H_b$, the human queue length $Q_H$, idle robot identities $\mathcal A_R$, current worker and robot locations, robot battery states, and the occupancy and queue length of each shared facility. The fleet size is $F=2$. Human delivery and robot support assignments both contribute to $Q_H$.

For facility $k\in\{E,C,D\}$, let $N_k$, $Q_k$, and $c_k$ denote occupied servers, waiting requests, and capacity. Here $E$, $C$, and $D$ denote the elevator bank, chargers, and cleaning stations. Define
\begin{align}
H_0(t)&=\max\{0,H_a(t)-H_b(t)\},\\
P_H(t)&=Q_H(t)/\max\{1,H_a(t)\},\\
P_E(t)&=\{N_E(t)+Q_E(t)\}/c_E,\\
P_S(t)&=\max\left\{\frac{N_C(t)+Q_C(t)}{c_C},
                       \frac{N_D(t)+Q_D(t)}{c_D}\right\}.
\end{align}
The station pressure $P_S$ concerns charging and cleaning. It is distinct from the availability of staff to perform a handoff, which is represented by the human queue. Battery states affect execution but are not an ARAD admission gate.

Let $D(u,v)$, $n_E(u,v)$, and $n_F(u,v)$ be the estimated route distance, elevator rides, and floors crossed. For mode $m\in\{H,R\}$, deterministic travel is
\begin{equation}
T_m(u,v)=\frac{D(u,v)}{v_m}+2.2n_E(u,v)+0.25n_F(u,v).
\label{updated-eq:route-prediction}
\end{equation}
This estimate includes nominal elevator service but excludes elevator queueing, random congestion, and handling. Observed delivery pair estimates take precedence in the router. Other pairs, including travel to pickups, use shortest paths in the directed pathway graph. The graph encodes distance and elevator transitions, not robot accessibility.

An idle robot is selected by its estimated travel to the pickup:
\begin{equation}
r_i^*(t)=\mathop{\arg\min}_{r\in\mathcal A_R(t)}T_R(\ell_r(t),o_i).
\label{updated-eq:nearest-robot}
\end{equation}
Robot identity resolves ties. For a human or robot starting immediately, the completion estimates are
\begin{align}
\widehat C_{ij}^{H}(t)&=t+T_H(\ell_j(t),o_i)+L_i
                         +T_H(o_i,d_i)+R_i,\\
\widehat C_i^{R}(t)&=t+T_R(\ell_{r_i^*}(t),o_i)+L_i
                         +T_R(o_i,d_i)+R_i.
\label{updated-eq:robot-prediction}
\end{align}
Locations are used in the primary model for both modes. The robot estimate refers to the selected idle unit, rather than the first unit in the fleet store. A robot assignment is made only at request arrival. Requests placed in the human queue are not reconsidered for a robot later.

The human dispatcher first chooses the waiting delivery or support assignment with the smallest tuple of priority, deadline, and insertion order. It then assigns that work to the available transporter with the smallest estimated travel to the required location. Worker identity resolves ties. This is an online priority rule with proximity based worker selection. It does not jointly optimize the queue or anticipate future requests.

\subsection{ARAD and the deadline admission benchmark}
\label{updated-sec:arad-gates}
Set $s_i(t)=\delta_i-t$ and $\widehat\tau_i^R(t)=\widehat C_i^R(t)-t$. The elevator adjusted signal is
\begin{equation}
\widetilde\tau_i^R(t)=\widehat\tau_i^R(t)\{1+\alpha P_E(t)\},
\qquad \alpha=0.15.
\label{updated-eq:elevator-signal}
\end{equation}
The inflation is a fixed rule parameter, not an estimate fitted to elevator observations. With $A(t)=|\mathcal A_R(t)|$, the common admission gate is
\begin{equation}
g_i^0(t)=\mathbf{1}\{A(t)>0\}\mathbf{1}\{H_0(t)=0\}
           \mathbf{1}\{\widetilde\tau_i^R(t)\leq s_i(t)\}.
\label{updated-eq:common-gate}
\end{equation}
If no robot is idle, the gate is zero without evaluating a robot route. The reserve is $R(F)=\min\{1,\max(0,F-1)\}$, which is one for the tested fleet. With $\rho_H=0.25$ and $\rho_S=1$, the additional nonurgent gate is
\begin{align}
g_i^N(t)=g_i^0(t)&\mathbf{1}\{P_H(t)\geq\rho_H\}\nonumber\\
{} \times&\mathbf{1}\{P_S(t)\leq\rho_S\}
\mathbf{1}\{A(t)>R(F)\}.
\end{align}
ARAD assigns a request to the selected robot when
\begin{equation}
y_i^{A}(t)=
\begin{cases}
g_i^0(t), & p_i=1,\\
g_i^N(t), & p_i>1.
\end{cases}
\label{updated-eq:arad-decision}
\end{equation}
Here $y_i=1$ denotes robot assignment; otherwise the request enters the human queue. Urgent work bypasses the nonurgent pressure and reserve conditions. ARAD is a deterministic map from observable state and request attributes to an assignment. The word adaptive refers to response to that state. Parameters are neither learned nor adjusted during a run, and the rule does not solve an optimization over future arrivals.

The simpler deadline admission rule assigns according to
\begin{equation}
y_i^{D}(t)=g_i^0(t)
\label{updated-eq:deadline-rule}
\end{equation}
for every priority. It retains the same nearest robot, pickup travel, elevator inflation, and human availability conditions. It removes the three additional conditions for nonurgent work together. This is a benchmark for the combined restriction, rather than an ablation identifying the effect of each individual condition. The name denotes its admission test. It does not guarantee completion before the deadline.

\subsection{Proximity and workload benchmark}
\label{updated-sec:proximity-rule}
The proximity benchmark compares a robot signal with a human signal. Let $\mathcal A_H(t)$ contain available transporters and define the direct human service estimate $\widehat\tau_i^{H,\mathrm{dir}}=L_i+T_H(o_i,d_i)+R_i$. Its human signal is
\begin{equation}
\widehat\tau_i^{H,P}(t)=
\begin{cases}
\displaystyle\min_{j\in\mathcal A_H(t)}
\{\widehat C_{ij}^H(t)-t\}, & \mathcal A_H(t)\ne\varnothing,\\
\widehat\tau_i^{H,\mathrm{dir}}
\left\{1.5+\dfrac{Q_H(t)+1}{\max(1,H_a(t))}\right\},
& \mathcal A_H(t)=\varnothing.
\end{cases}
\label{updated-eq:human-proxy}
\end{equation}
The second branch is a workload proxy. It does not predict the future location or remaining service time of a busy worker. The added support proxy is
\begin{equation}
\widehat S_i(t)=(L_i+R_i+2)\{1+P_H(t)\}.
\label{updated-eq:support-proxy}
\end{equation}
The robot is admitted when
\begin{align}
y_i^P(t)=\mathbf{1}\{A(t)>0\}
&\mathbf{1}\{\widehat\tau_i^R(t)\leq s_i(t)\}\nonumber\\
{}\times&\mathbf{1}\{\widehat\tau_i^R(t)+\widehat S_i(t)
                       <\widehat\tau_i^{H,P}(t)\}.
\label{updated-eq:proximity-decision}
\end{align}
The added proxy deliberately penalizes human support even though loading and receiving are already present in $\widehat\tau_i^R$. It is not a calibrated completion predictor. It excludes cleaning and explicit failure recovery. Unlike ARAD, this rule can select a robot while a human is idle if the comparison favors the robot. It represents a transparent greedy comparison of estimated service, related to task allocation concepts \cite{gerkey2004}, rather than a reproduction of a published scheduler.

\subsection{Admission, completion, and robot availability}
\label{updated-sec:mission-execution}
Execution uses the same resource model for every robot policy. After selection, the robot acquires its unit and draws planned travel times. Required energy is $e_i=0.025(\tau_i^{\mathrm{pickup}}+\tau_i^{\mathrm{delivery}}+L_i+R_i)$ kWh, where the travel terms include the congestion draw and nominal elevator service. If the battery cannot cover $e_i$, the selected unit charges before beginning the mission. A mission exceeding full battery capacity raises an infeasibility error. No such event occurred in the reported grids. Elevator queueing, support waiting, cleaning, standby, and recovery do not contribute to this energy estimate.

The robot reaches the origin and requests loading, coordination, and any recovery from the transporter queue. Coordination is $2\exp(Z)$ minutes, with $Z\sim N(0,0.3^2)$. Recovery is added after a modeled failure and uses an exponential duration with mean 12 minutes. Receiving creates a second human assignment at the destination. Each support assignment requires the selected worker to reach its handoff location and occupies that worker until support finishes.

The resulting completion time can be written as
\begin{align}
C_i^R=t+W_i^{\mathrm{charge}}+\tau_i^{\mathrm{pickup}}
&+\Gamma_{i,o}^{H}+L_i+K_i+F_i\nonumber\\
&+\tau_i^{\mathrm{delivery}}+\Gamma_{i,d}^{H}+R_i,
\label{updated-eq:realized-completion}
\end{align}
where realized travel includes elevator queueing, $K_i$ is coordination, $F_i$ is recovery when needed, and $\Gamma_{i,o}^{H}$ and $\Gamma_{i,d}^{H}$ include queueing and worker travel before the corresponding support action. $W_i^{\mathrm{charge}}$ includes any initial charging wait and service. These terms are not all represented in the admission signals. A feasible signal cannot establish a feasible completed mission.

Delivery completion is recorded after receiving. Cleaning follows completion for specimens and instruments. It first requests two minutes of human support at the destination, then occupies a cleaning station for seven minutes. This work affects the availability of robots and staff for subsequent requests, although it is not added to the completed request's lateness. A robot at or below 25\% charge then enters charging; otherwise it returns to the idle fleet. Distinguishing admission, delivery completion, and return to availability is essential when comparing controllers that admit different numbers of missions.

\subsection{Outcomes and inference}
The main outcome is the fraction of requests created during measurement that finish after their assumed deadline or remain incomplete after a one day clearance period. Urgent lateness applies the same definition to priority 1. We also report mean time from request to completion, human empty travel, human support minutes, contiguous idle intervals of at least ten minutes, robot completions, and utilization. Empty travel includes movement to delivery pickups and support locations. Support travel is recorded separately as a subset. Utilization counts delivery, pickup travel, and support, rather than treating support as idle. Aggregate utilization is a measure of occupied shift time, not a measure of whether a worker is available at the needed location.

Let $\mathcal I_j$ contain the idle intervals of worker $j$ within measured shifts, after removing all delivery, pickup travel, support travel, and support service. The interval measure is
\begin{equation}
I_\tau=\frac{100}{BD}\sum_j\sum_{\ell\in\mathcal I_j}
                 \ell\mathbf{1}\{\ell\geq\tau\},
\qquad \tau=10\ \mathrm{min}.
\label{updated-eq:idle-intervals}
\end{equation}
Here $B=200$ is the nominal bed denominator and $D=30$ is the number of measured days. The entire qualifying interval is counted. Support occupies the worker's schedule and can split an interval; it is not subtracted as aggregate minutes after the run. The threshold is an analytical convention, not a validated duration for clinical work. Robot coverage is the mean across replications of robot completions divided by requests created during measurement.

For an outcome $Y$, each hybrid policy is compared with the human only policy using the same demand seed. Direct ARAD minus proximity contrasts also pair the two policies within each seed. With $n=30$ replications, the pointwise Monte Carlo interval for the paired mean is $\bar d\pm t_{0.975,29}s_d/\sqrt{30}$. The interval measures simulation variation under fixed assumptions. It does not cover missing input data, structural error, or a hospital population. There is no multiplicity correction, so comparisons across scenarios are descriptive.

Each full grid has two rosters, three specimen deadlines, two task mixes, and three policies, for 36 combinations at a fixed elevator capacity. Every confirmation run has seven warmup days and 30 measured days. Screening used five shorter replications at the initial capacity and was not used for inferential claims. The four server grid was added after the initial result exposed a bottleneck, so its comparisons are exploratory. The additional deadline benchmark covers the source count mix and 25 minute specimen deadline at both rosters and both elevator capacities, giving four scenarios and 120 further runs. It uses the original confirmation seeds 20261120 through 20261149. Its parameters are inherited from the earlier implementation without tuning to these outcomes. This extension and all direct policy contrasts were added after the main grids and are exploratory.

Random operational draws can diverge after dispatch choices, so pairing preserves the demand seed but does not synchronize every service draw. Each policy pair is checked for matching request counts and urgent request counts before analysis. Reporting follows the STRESS guidance for simulation studies \cite{monks2019}. A larger number of replications reduces Monte Carlo error but cannot establish which assumed staffing or elevator configuration represents actual service.

\section{Results of the Updated Operational Study}

\subsection{Effect of representing pickup travel}
The matched human only ablation is revealing. With the source count mix and a 25 minute specimen deadline, omitting pickup travel gives 5.25\% late or incomplete requests with the high roster and 5.26\% with the moderate roster. Adding pathway travel between jobs raises the mean to 55.11\% with the high roster and 72.76\% with the moderate roster. Across the same 30 demand seeds, the paired increases are 49.85 percentage points (95\% interval, 48.48 to 51.22) and 67.50 points (66.28 to 68.73), respectively. These are differences between two assumed operating models, not estimates of actual hospital lateness. Mean modeled elevator wait in the high roster changes from 4.01 to 36.76 minutes per elevator request. In the moderate roster it changes from 3.98 to 24.50 minutes. Pickup travel both occupies transporters and adds trips to the shared elevator bank. The high roster has shorter human queue delays, but its prompt assignment of more trips can create greater elevator pressure.

\begin{figure}[t]
\centering
\includegraphics[width=\columnwidth]{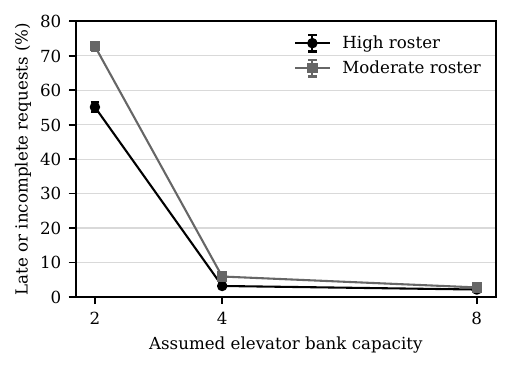}
\caption{Human only lateness under three assumed pooled elevator capacities. The source count mix and 25 minute specimen deadline are fixed. Bars show pointwise 95\% Monte Carlo intervals over 30 replications.}
\label{updated-fig:elevator-sensitivity}
\end{figure}

Fig.~\ref{updated-fig:elevator-sensitivity} shows how strongly this result depends on the assumed elevator bank. With four servers and the same source count mix and 25 minute deadline, human only lateness is 3.18\% for the high roster and 5.92\% for the moderate roster. The paired changes from the two server scenario are $-51.93$ points (95\% interval, $-53.51$ to $-50.35$) and $-66.84$ points ($-68.15$ to $-65.53$). With eight servers, lateness is 2.14\% and 2.75\%. Mean elevator wait falls to 0.50 and 0.90 minute at four servers, then 0.01 and 0.02 minute at eight. Thus, the high late fractions at two servers are primarily a consequence of the modeled pooled elevator bottleneck, amplified by pickup trips. The actual capacity and passenger load were not observed.

\subsection{Policy comparisons}
Table~\ref{updated-tab:operational-main} reports the entire confirmed scenario grid. The high roster with the source count mix and a 25 minute specimen deadline has 55.11\% late or incomplete requests under human only dispatch. Relative to that paired baseline, ARAD changes lateness by $+0.04$ percentage point, with a pointwise 95\% Monte Carlo interval from $-0.07$ to $+0.15$. The proximity rule changes it by $+0.24$ point, with an interval from $+0.10$ to $+0.38$. Under the moderate roster, the corresponding human baseline is 72.76\%, and the increases are $+0.22$ point ($+0.15$ to $+0.29$) and $+0.43$ point ($+0.35$ to $+0.51$). These intervals describe simulation variation, not clinical uncertainty.

\begin{table*}[t]
\caption{Overall late or incomplete requests with two assumed elevator servers in 30 replications. Differences are percentage points relative to the paired human only case.}
\label{updated-tab:operational-main}
\centering\footnotesize
\begin{tabular}{lllrrr}
\toprule
Roster & Task mix & Specimen deadline (min) & Human only (\%) & ARAD difference & Proximity difference \\
\midrule
High & Source & 25 & 55.11 & +0.04 & +0.24 \\
High & Source & 45 & 43.82 & +0.19 & +0.21 \\
High & Source & 60 & 38.26 & +0.31 & +0.23 \\
\addlinespace
High & Balanced & 25 & 46.27 & -0.16 & +0.01 \\
High & Balanced & 45 & 41.04 & +0.01 & -0.06 \\
High & Balanced & 60 & 38.44 & -0.10 & -0.09 \\
\midrule
Moderate & Source & 25 & 72.76 & +0.22 & +0.43 \\
Moderate & Source & 45 & 66.44 & +0.56 & +0.39 \\
Moderate & Source & 60 & 63.39 & +0.69 & +0.46 \\
\addlinespace
Moderate & Balanced & 25 & 67.39 & +0.25 & +0.32 \\
Moderate & Balanced & 45 & 63.70 & +0.81 & +0.46 \\
Moderate & Balanced & 60 & 60.33 & +1.36 & +0.56 \\
\bottomrule
\end{tabular}
\end{table*}

\begin{figure*}[t]
\centering
\includegraphics[width=0.93\textwidth]{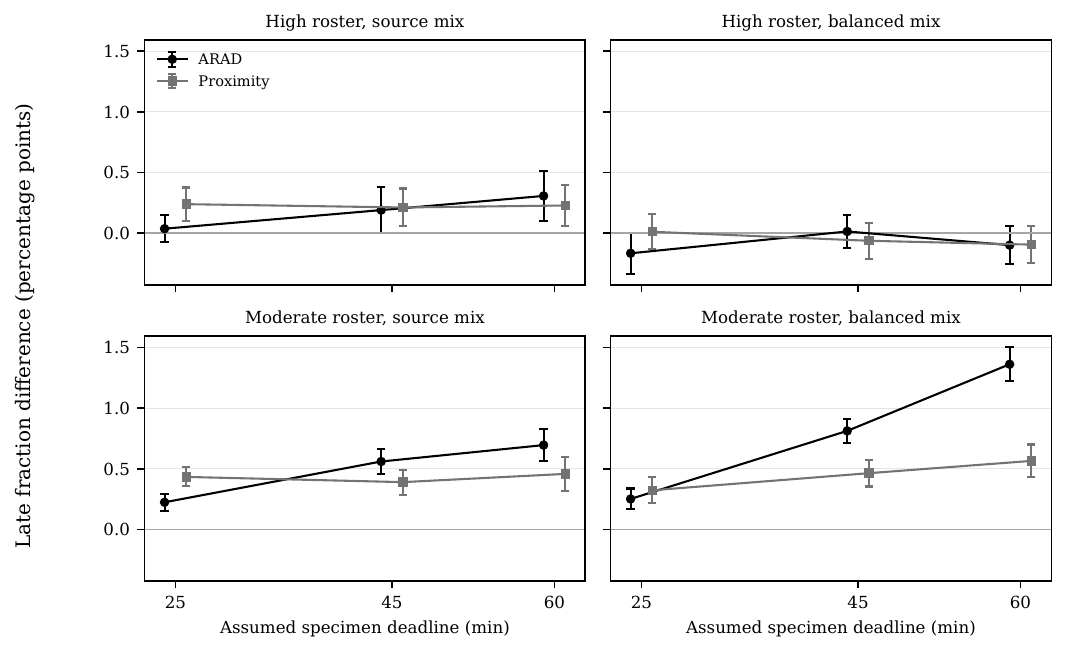}
\caption{Hybrid policy minus human only late or incomplete fraction, in percentage points. Points are means; bars are pointwise 95\% Monte Carlo intervals over 30 paired replications. Task mix, deadline, and roster are assumed scenarios.}
\label{updated-fig:operational-policy}
\end{figure*}

Fig.~\ref{updated-fig:operational-policy} shows the paired policy effects and their pointwise intervals. Across the 12 operating scenarios, ARAD's mean difference ranges from $-0.16$ to $+1.36$ percentage points, while the proximity rule ranges from $-0.09$ to $+0.56$. Every mean difference under moderate staffing is positive. Some high roster contrasts are close to zero or slightly negative, but none shows a substantial stable gain. The policies admit relatively few robot missions. In the source count, 25 minute case, ARAD completes a mean of 16.8 robot requests over 30 measured days with the high roster and 90.6 with the moderate roster; the proximity rule completes 46.0 and 131.4. The paired demand averages 11,025 requests in that interval. These admission rules leave the overwhelming majority of deliveries with humans.

\subsection{Policy comparisons with four elevator servers}
The less congested bank changes absolute service but does not produce a robot benefit. Table~\ref{updated-tab:operational-main-four} reports the second full grid. In the high roster, source count mix, 25 minute case, the human only late fraction is 3.18\%. ARAD adds $0.11$ percentage point (95\% interval, $0.05$ to $0.18$), while proximity adds $0.03$ point ($-0.02$ to $0.08$). In the moderate roster, human only lateness is 5.92\%; ARAD adds $1.28$ points ($1.08$ to $1.49$) and proximity adds $0.08$ point ($0.00$ to $0.15$). The moderate roster's urgent late fraction is 6.78\% under human only dispatch. ARAD raises it by $1.51$ points ($1.27$ to $1.74$). These are exploratory pointwise intervals because the four server grid followed inspection of the two server result.

\begin{table*}[t]
\caption{Overall late or incomplete requests with four assumed elevator servers in 30 replications. Differences are percentage points relative to the paired human only case.}
\label{updated-tab:operational-main-four}
\centering\footnotesize
\begin{tabular}{lllrrr}
\toprule
Roster & Task mix & Specimen deadline (min) & Human only (\%) & ARAD difference & Proximity difference \\
\midrule
High & Source & 25 & 3.18 & +0.11 & +0.03 \\
High & Source & 45 & 0.22 & +0.03 & -0.00 \\
High & Source & 60 & 0.16 & +0.01 & +0.00 \\
\addlinespace
High & Balanced & 25 & 1.61 & +0.05 & +0.00 \\
High & Balanced & 45 & 0.25 & +0.01 & +0.00 \\
High & Balanced & 60 & 0.22 & +0.00 & -0.00 \\
\midrule
Moderate & Source & 25 & 5.92 & +1.28 & +0.08 \\
Moderate & Source & 45 & 0.52 & +0.34 & +0.04 \\
Moderate & Source & 60 & 0.33 & +0.10 & +0.03 \\
\addlinespace
Moderate & Balanced & 25 & 3.03 & +0.88 & +0.11 \\
Moderate & Balanced & 45 & 0.56 & +0.26 & +0.06 \\
Moderate & Balanced & 60 & 0.51 & +0.08 & +0.05 \\
\bottomrule
\end{tabular}
\end{table*}

\begin{figure*}[t]
\centering
\includegraphics[width=0.93\textwidth]{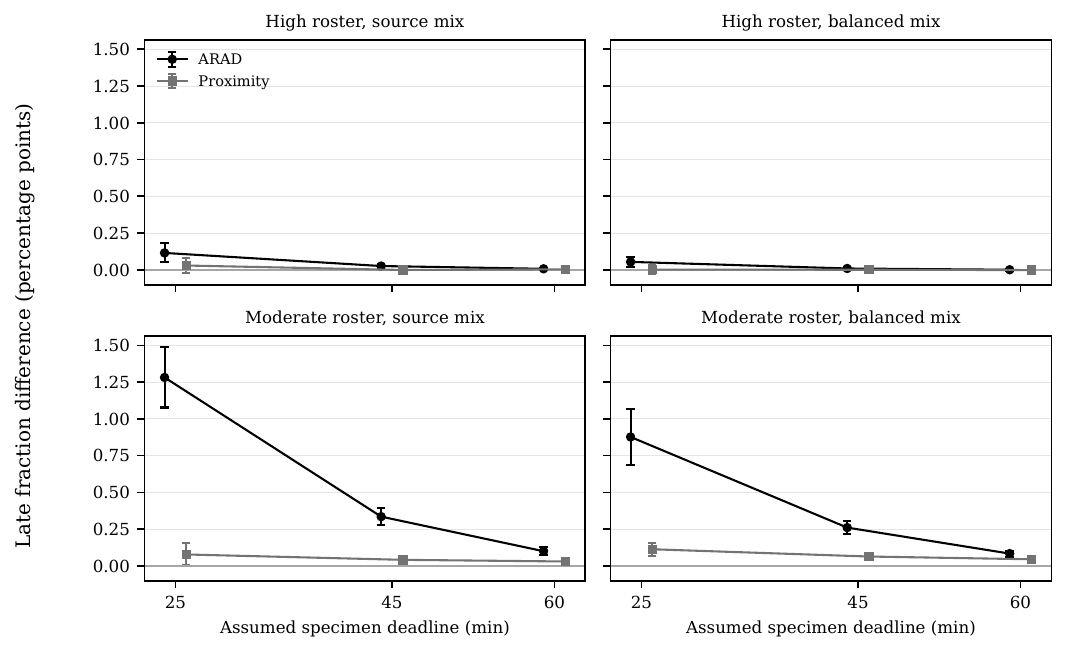}
\caption{Hybrid policy minus human only lateness under four assumed elevator servers. Points are means; bars are pointwise 95\% Monte Carlo intervals over 30 paired replications.}
\label{updated-fig:operational-policy-four}
\end{figure*}

Fig.~\ref{updated-fig:operational-policy-four} shows that the four server ARAD differences are predominantly positive. With the moderate roster and source count mix, ARAD completes a mean of 274.6 robot requests over 30 days and requires 28.1 support minutes per 100 nominal beds per day. The proximity rule completes 35.9 requests and requires 3.5 support minutes. These counts explain part of their different effects. The comparator is more selective, and neither rule substitutes for a policy optimized under measured operational costs.

\subsection{Deadline and task composition}
The assumed specimen deadline changes outcomes much more than the robot rule. With the high roster and source count mix, human only lateness falls from 55.11\% at 25 minutes to 38.26\% at 60 minutes. Under the balanced mix it falls from 46.27\% to 38.44\%. With the moderate roster and source count mix, it falls from 72.76\% to 63.39\%. With four servers, the high source count case falls from 3.18\% to 0.16\%, and the moderate case falls from 5.92\% to 0.33\%. These contrasts do not identify the true deadline. They show the consequence of assigning one to an event stream without observed urgency.

With two servers, human empty travel in the source count, 25 minute human only case averages 3,985 and 4,002 minutes per 100 nominal beds per day for the high and moderate rosters. That measure includes time spent reaching delivery pickups, not just distance traveled. Human utilization is 0.32 and 0.57. The high late fraction despite low utilization shows that a resource can be unavailable at the relevant place or blocked at an elevator while aggregate staffed hours remain. Robot handoff work remains small because the rules admit few missions. ARAD adds 1.7 and 8.9 support minutes per 100 nominal beds per day in the high and moderate cases. The proximity rule adds 4.6 and 12.8.

\subsection{Direct policy contrasts and the deadline benchmark}

\label{updated-sec:benchmark-results}

Table~\ref{updated-tab:reconciled-benchmarks} adds the simpler deadline admission rule to four representative cases, using the same source count mix, 25 minute deadline, rosters, and elevator capacities as the main grids. The additional 120 runs use the same 30 demand seeds and the same pickup travel, support, charging, and cleaning model. These comparisons are exploratory. They do not reuse the older experiment without pickup travel.

\begin{table*}[t]
\caption{Exploratory benchmark comparisons for the source count mix and 25 minute specimen deadline. Bank denotes assumed elevator servers. Every contrast pairs 30 runs by demand seed against human only dispatch. Late differences are percentage points; idle differences and support are minutes per 100 nominal beds per day. Brackets give pointwise 95\% Monte Carlo intervals. Coverage is the mean fraction of requests completed by robots. Support excludes worker travel, which is recorded separately.}
\label{updated-tab:reconciled-benchmarks}
\centering\footnotesize
\setlength{\tabcolsep}{4pt}
\begin{tabular}{lllrlllr}
\toprule
Bank & Roster & Policy & Coverage (\%) & Overall difference & Urgent difference & Idle difference & Support \\
\midrule
2 & High & ARAD & 0.15 & \shortstack{+0.04\\{}[-0.07, +0.15]} & \shortstack{+0.03\\{}[-0.09, +0.14]} & \shortstack{-8.37\\{}[-32.44, +15.70]} & 1.7 \\
2 & High & Proximity & 0.42 & \shortstack{+0.24\\{}[+0.10, +0.38]} & \shortstack{+0.22\\{}[+0.09, +0.36]} & \shortstack{-60.35\\{}[-90.62, -30.09]} & 4.6 \\
2 & High & Deadline & 0.19 & \shortstack{+0.06\\{}[-0.09, +0.22]} & \shortstack{+0.02\\{}[-0.14, +0.19]} & \shortstack{-18.95\\{}[-44.71, +6.80]} & 2.1 \\
\addlinespace
2 & Moderate & ARAD & 0.82 & \shortstack{+0.22\\{}[+0.15, +0.29]} & \shortstack{+0.23\\{}[+0.16, +0.30]} & \shortstack{-37.24\\{}[-45.80, -28.67]} & 8.9 \\
2 & Moderate & Proximity & 1.19 & \shortstack{+0.43\\{}[+0.35, +0.51]} & \shortstack{+0.43\\{}[+0.34, +0.52]} & \shortstack{-74.20\\{}[-83.91, -64.48]} & 12.8 \\
2 & Moderate & Deadline & 0.84 & \shortstack{+0.22\\{}[+0.16, +0.29]} & \shortstack{+0.20\\{}[+0.13, +0.27]} & \shortstack{-32.09\\{}[-39.83, -24.35]} & 8.8 \\
\addlinespace
4 & High & ARAD & 0.34 & \shortstack{+0.11\\{}[+0.05, +0.18]} & \shortstack{+0.14\\{}[+0.06, +0.21]} & \shortstack{-0.24\\{}[-1.05, +0.57]} & 3.9 \\
4 & High & Proximity & 0.11 & \shortstack{+0.03\\{}[-0.02, +0.08]} & \shortstack{+0.03\\{}[-0.03, +0.09]} & \shortstack{-0.19\\{}[-2.11, +1.73]} & 1.2 \\
4 & High & Deadline & 0.41 & \shortstack{+0.12\\{}[+0.05, +0.18]} & \shortstack{+0.13\\{}[+0.06, +0.19]} & \shortstack{+0.56\\{}[-0.46, +1.59]} & 4.5 \\
\addlinespace
4 & Moderate & ARAD & 2.48 & \shortstack{+1.28\\{}[+1.08, +1.49]} & \shortstack{+1.51\\{}[+1.27, +1.74]} & \shortstack{-11.43\\{}[-14.37, -8.49]} & 28.1 \\
4 & Moderate & Proximity & 0.33 & \shortstack{+0.08\\{}[0.00, +0.15]} & \shortstack{+0.08\\{}[0.00, +0.16]} & \shortstack{-0.68\\{}[-3.69, +2.32]} & 3.5 \\
4 & Moderate & Deadline & 2.75 & \shortstack{+1.30\\{}[+1.10, +1.50]} & \shortstack{+1.36\\{}[+1.14, +1.58]} & \shortstack{-9.65\\{}[-13.18, -6.12]} & 29.9 \\
\addlinespace
\bottomrule
\end{tabular}
\end{table*}

With 2 elevator servers, the paired ARAD minus proximity difference in overall lateness is -0.20 point (-0.36 to -0.04) with the high roster and -0.21 point (-0.29 to -0.13) with the moderate roster. The intervals are calculated from direct differences within each seed.

With 4 elevator servers, the paired ARAD minus proximity difference in overall lateness is +0.09 point (+0.01 to +0.17) with the high roster and +1.20 point (+0.99 to +1.42) with the moderate roster. The intervals are calculated from direct differences within each seed.

Across all 24 scenarios in the two main grids, the ARAD minus proximity mean difference ranges from $-0.21$ to $+1.20$ percentage points. Pointwise intervals lie below zero in 2 scenarios and above zero in 13; the other 9 cross zero. This is a descriptive summary of exploratory contrasts, without multiplicity correction. It does not identify one policy as uniformly preferable.

Removing the three nonurgent ARAD conditions together changes overall lateness relative to ARAD by $+0.03$ point (-0.13 to +0.18) with the high roster and $0.00$ point (-0.07 to +0.07) with the moderate roster at 2 servers. This contrast concerns the combined admission restriction. It cannot attribute an effect to an individual gate.

Removing the three nonurgent ARAD conditions together changes overall lateness relative to ARAD by $0.00$ point (-0.02 to +0.03) with the high roster and $+0.02$ point (-0.14 to +0.17) with the moderate roster at 4 servers. This contrast concerns the combined admission restriction. It cannot attribute an effect to an individual gate.

The deadline and ARAD overall contrasts all have intervals crossing zero in these four cases. This limited separation is consistent with the scenario composition: specimens account for 84.06\% of requests and have priority 1, so both rules use the common gate for most requests. The extension does not establish that the nonurgent gates are unimportant under a different task mix.

Coverage and support help interpret the service differences. A controller that admits more robot requests also schedules more handoff work on the transporter queue. The benchmark comparisons evaluate complete workflows, rather than matched numbers of robot missions. Coverage is descriptive; dividing the overall policy effect by robot completions would not estimate a causal effect per robot trip, because assignments change subsequent queues and resource states.

\subsection{Verification of the updated calculations}

All 2,160 runs in the two full grids and the additional 120 deadline benchmark runs satisfy the conservation identity: measured requests created equal completed plus incomplete requests. No robot mission exceeded full battery capacity. The two original grids recorded one and seven forced charges, respectively. The benchmark extension uses the same simulator, route implementation, configurations, and processed inputs. Saved hashes confirm the provenance of the original raw files before the extension is analyzed.

Eighteen simulation and routing tests passed for the reported implementation. They cover request accounting, reproducibility, pickup routes, support and travel accounting, robot charging, and existing dispatch behavior. Direct policy comparisons additionally check that every pair has the same seed, scenario attributes, request count, and urgent request count. Their intervals are derived from the paired raw outcomes, rather than from differences between summary intervals.

These checks establish internal consistency and traceability. The model has not been checked against observed delivery completion times or staff movements. The distinction matters: conservation can hold in a model whose service assumptions are inappropriate for deployment. The benchmark extension improves the comparison within the implemented model without establishing external validity.

\section{Verification of Earlier Experiments}
\input{verification_results.tex}

The revised simulation generated 820 experiment rows and 450 staffing confirmation rows. Conservation checks matched created, completed, and incomplete requests. Unit tests cover support queueing, empirical route graph fallback, sampled priorities, repositioning, battery admission, and passenger elevator traffic. No primary mission violated battery feasibility. These checks establish internal properties, not clinical validity.

Source, configurations, processed inputs, raw results, and generated tables are retained locally with SHA-256 checksums. The reanalysis in \path{scripts/review_publication_analysis.py} did not change simulation outcomes. Empirical route lengths support the network's physical scale, but completion times and actual staffing were unavailable for external validation.

\section{Discussion of Earlier Experiments}
\subsection{What the primary comparison supports}
The revised model does not support the earlier favorable robot result. Once the human roster is recalibrated and robot support competes for the same workers, ARAD and deadline feasible dispatch both worsen lateness. They also reduce long idle intervals. More robots increase support demand and further degrade urgent service. Within these assumptions, moving a journey to a robot changes the form and timing of human work but does not release capacity.

This reversal is more informative than an editorial qualification. The earlier component model counted coordination and recovery only as aggregate burden. It could subtract minutes after the run, but it could not represent a transporter waiting to support a robot, a delayed handoff, or the fragmentation of an otherwise usable idle interval. The revised queue captures those interactions. Its result does not prove that hospital delivery robots are ineffective. It shows that a benefit claim depends on who performs each handoff, when that work occurs, and whether the human comparator is already feasible.

The deadline feasible benchmark and ablations do not reveal a hidden advantage for the ARAD gates. Removing individual gates changes overall lateness by at most about 0.13 percentage points in the tested ablation set. The support pressure gate never binds. The formalization is nevertheless useful because it identifies which state variables affect decisions, which gates are inactive, and where a future learned or optimization based controller would need better observations. The main empirical contrast remains between human only service and a shared human robot workflow, not between minor variations of the rule.

\subsection{Implications for hospital robot deployment}
The passenger scenarios show that building operations can dominate dispatch refinements. Six assumed passenger elevator requests per hour more than doubled lateness in both the human and robot cases. This does not estimate Freiburg passenger traffic, but it shows that a robot evaluation based only on freight demand can be badly misleading. Elevator logs should include passenger arrivals, direction, boarding delays, capacity, and service by time of day.

The repositioning stress result is similarly consequential. Adding travel from the previous destination to the next pickup without changing the central priority rule caused lateness above 92 percent. The result is too structurally crude for procurement, because real dispatch would consider worker location, standby points, zones, and task compatibility. It nevertheless shows that zero pickup travel is not a harmless convenience. A credible next model needs observed trajectories or location events and a nearest feasible assignment benchmark.

The task scenarios also show why the source event counts cannot be treated as a measured workload. The balanced mix greatly reduces lateness, and mixed priorities change urgent results and relative policy behavior. Hospitals considering deployment should collect one record per physical transport, with contents, urgency, request time, pickup readiness, origin, destination, assignment, completion, and every human interaction. These data would support calibration and external validation rather than scenario interpretation alone.

\subsection{Answering the capacity question}
The revised primary result answers the modeled capacity question negatively. The tested robots do not produce longer usable transporter intervals, and their support work slightly worsens deadlines. This is a statement about the modeled workflow, not about nursing capacity or all hospital robots. Human resources in the simulation are transporters. No inference can be made about care delivered, patient outcomes, or the value of reducing exposure to hazardous loads.

A different workflow could change the result. Dedicated loading staff, secure automated handoff, route consolidation, off peak scheduling, or a robot fleet serving tasks with long deadlines could reduce support interference. Such designs should be evaluated as complete operating models, with their staffing and costs included. Minutes of nominal travel avoided are not an adequate substitute for that analysis.

\subsection{Limitations and required validation}
Specimens account for 84.06 percent of the source mapped scenario and every specimen has priority 1. Laboratory rows can represent several assays from one collection, while prescriptions and dispensing events do not map one to one to trips. Instrument and supply proxies are especially weak. Random class assignment ignores clinical route constraints. Sensitivity analysis demonstrates the effect of these choices but cannot identify the correct mix.

Support durations are assumed rather than observed. The same generic transporter performs delivery and robot support, and support inherits the task priority and deadline. Breaks, competing duties, recipient availability, secure compartment access, and role restrictions are absent. Cleaning is represented as fixed downtime plus two minutes of human work rather than a validated infection control process.

The primary model still omits pickup travel. The repositioning variant adds it to a dispatch rule that ignores proximity, so it is a stress test rather than a correction. The elevator model aggregates route elevator service into one bank request. Passenger streams are assumed, not measured, and individual cabins, direction, floor queues, and boarding capacity are absent. Energy parameters are engineering scenarios rather than measurements from a selected robot.

The staffing criteria, 200 bed denominator, deadlines, and service targets are analytical choices. The model has not been calibrated against observed completion times or actual Freiburg staffing. No independent hospital dataset was available for external validation. Monte Carlo intervals quantify simulation error under fixed assumptions and exclude structural and input uncertainty. The evidence does not establish infection prevention, clinical safety, financial return, additional care, or released nursing time.

\Needspace{9\baselineskip}
\subsection{Conclusion from the earlier experiments}
We formulated ARAD as a reproducible state feedback controller and embedded it in a coupled model of robot missions, transporter support, elevators, charging, and cleaning. After recalibrating the human roster and scheduling robot support on the same workers, the tested robot fleets increased lateness and reduced long idle intervals. A deadline feasible benchmark and gate ablations did not change that conclusion. Task mapping, passenger elevator traffic, and movement to the next pickup had larger effects than small dispatch variations. The formal decision model improves traceability, but it does not create evidence of a capacity gain. A deployment claim requires observed transport contents and deadlines, measured support and pickup travel, location aware robot selection, building traffic, and external validation in the intended hospital workflow.

\section{Discussion of the Updated Operational Study}
The updated analysis incorporates travel and worker selection throughout the main operating model. It gives a more complete comparison than the earlier primary experiment or its separate repositioning stress case. The following interpretation applies to the updated scenario grids and deadline benchmark.

\subsection{What the policy comparison establishes}
The operational conclusion is governed by model structure and uncertain inputs. A delivery robot can be admitted because its route appears feasible, then wait for a transporter to load or receive it. The same transporter may first cross the campus to reach that handoff. The added work is small per mission but arrives at times that compete with other requests. The limited number of robot completions also means that small differences between robot rules cannot be read as a robust fleet benefit.

The benchmark rules answer different questions. Human only dispatch tests the same location aware workforce without robot handoffs. Proximity dispatch uses a relative service signal and can favor a robot even when a human is idle. The deadline rule retains the common ARAD admission test while removing its extra nonurgent restrictions. Direct contrasts strengthen the interpretation of the admission logic. They do not establish superiority over an optimized scheduler. The policies also admit different numbers of missions, so the resulting service differences include their effects on support demand and future resource availability.

\subsection{Lessons for evaluating automation workflows}
The separation between admission and completion is useful beyond this hospital request stream. Define the prediction error $\epsilon_i=C_i^R-\widehat C_i^R$ and the nominal deadline margin $b_i=\delta_i-\widehat C_i^R$. For a completed request, the identity
\begin{equation}
C_i^R>\delta_i\quad\Longleftrightarrow\quad\epsilon_i>b_i
\label{updated-eq:admission-gap}
\end{equation}
shows why a nonnegative admission margin alone cannot establish timely service. In this model, prediction error contains charging, worker travel, support queueing, coordination, recovery, and deviations in travel time. Elevator inflation responds to one aggregate pressure signal but does not bound all these terms. A future controller could estimate their joint effect from operational records, while the current rules make their omission inspectable.

The same issue arises when transport work is evaluated as released staff time. A robot can remove a journey and introduce several handoffs at different times and locations. Total minutes do not preserve interval structure. For example, a two minute support task in the middle of a twenty minute idle interval can leave two nine minute intervals, neither of which meets a ten minute threshold. The current outcome accounts for such fragmentation directly through occupied worker schedules. This example explains the measure; it does not establish that ten minutes is useful for clinical care.

Shared infrastructure also changes the interpretation of staffing. In the two server source count case, high staffing coexists with substantial lateness and low aggregate human utilization. Additional staff do not remove an elevator queue, and prompt pickup assignments can increase the number of trips competing for the bank. This supports checking resource interactions before treating a dispatch refinement as the main cause of service performance. The precise sensitivity to two or four servers remains a property of the assumed pooled bank.

Together, these observations suggest a transferable evaluation practice: represent the complete alternative workflow, report robot exposure and scheduled human support, and distinguish service completion from the next opportunity to assign the vehicle. The numerical effects in this paper cannot be transferred to another building. The definitions and paired comparisons can be reused to examine whether a proposed automation result survives a more complete operating model.

\subsection{Model credibility and limits of transfer}

The large change in lateness when pickup travel is represented is an internal model comparison, not a measured deterioration at Freiburg. The matched ablation begins each human delivery at its pickup while retaining the same roster and demand. It omits relocation between deliveries from the workforce schedule. Yet the two server result is itself fragile: doubling assumed pooled elevator capacity removes most of its lateness. The capacity parameter cannot be inferred from the released request file or from the count of elevator edges in the graph. Even the current model is not a forecast. All workers start at one assumed node, and we have no observed locations, breaks, completion timestamps, task contents, or support durations with which to check the resulting travel and cycle times. We cannot determine which scenario is close to actual service performance.

The setting differs from published robot scheduling studies in ways that prevent a direct numerical reproduction. Published models use their own task definitions, time windows, hospital layouts, and support workflows \cite{cheng2023,ge2025,dei2026}. Here the observed request stream has no time windows or completion labels, and adding invented values to match a reported percentage would hide the uncertainty. The human only, proximity, and deadline rules give internal numerical reference points with the same arrivals and assumed operations. A comparison with a published optimization method would require an explicit adaptation of its objectives, information, and support constraints. It remains an important extension, rather than evidence supplied by the present benchmarks.

There are also physical omissions. The graph encodes pathway distance but not pedestrian density, door access, payload restrictions, or whether every route is robot accessible. One bank represents all elevator requests and omits passenger traffic. Energy excludes standby and some support movements. Cleaning is fixed downtime, not an infection prevention protocol. The model assumes every task class can be assigned to a robot. The synthetic MIMIC event mix has no known relationship to the fraction of real specimen transport. Pointwise intervals cannot repair these input and structural gaps.

For a deployment study, the next data collection should identify each physical transport, contents, priority, due time, pickup readiness, pickup and completion timestamps, worker or robot identifier, worker position before assignment, and all handoff work. Elevator arrivals and access delays should be logged in the intended building. Those observations would permit calibration of service times and a held out validation of lateness and worker movement. Only then would it be reasonable to optimize fleet size or claim released staff capacity.

\Needspace{9\baselineskip}
\section{Conclusions}

A coupled simulation of human travel, robot support, and shared facilities changes the interpretation of hospital robot dispatch. The formal controller specification separates the admission signal from realized completion and return to availability. Direct paired comparisons, a simpler deadline benchmark, and robot coverage make the policy evaluation more explicit. The tested rules do not show a stable service gain across the main scenario grids. Their effects are smaller than the interaction between pickup travel and assumed elevator capacity, alongside sensitivity to unobserved deadlines and task composition. The broader lesson is that dispatch should be evaluated against the complete human and robot workflow, including support travel and the timing of shared work. The code and scenario outputs permit reproduction of these conditional comparisons. They do not establish that robots improve, or fail to improve, actual Freiburg service.

\section*{Data and Code Availability}
Source datasets are available through cited persistent identifiers. Code and outputs are retained in the GitHub system \url{https://github.com/siwekk/medical_transport_robots}. The operational study also retains both full confirmation grids, the matched pickup travel and elevator comparisons, the deadline benchmark extension, and the paired analysis scripts.

\section*{Ethics Statement}
This simulation used public deidentified data and synthetic attributes. It did not recruit participants or intervene in care. No institutional ethics determination is reported. Statements about exemption or approval must reflect the authors' institutional requirements.

\section*{Submission Declarations}
The authors are Krzysztof Siwek, Warsaw University of Technology, and Aleksandra \ifmmode\acute{S}\else\'{S}\fi{}wietlicka, Pozna\'n University of Technology. 
%Author contributions, funding, 

\section*{Declaration of generative AI and AI-assisted technologies in the writing process}
During the preparation of this work, the authors used OpenAI Codex to assist with software development and Grammarly for language editing. The authors reviewed and edited all generated material, verified the calculations and citations, and take full responsibility for the content of the publication.

\input{supplement.tex}

\bibliographystyle{unsrt}
\bibliography{references}
\end{document}

%% file: literature_review.tex
\section{Related Work}
\label{sec:literature}

\subsection{Hospital logistics robots and simulation}

Hospital robotics includes material transport, disinfection, telepresence, rehabilitation, and direct clinical assistance. Reviews of this field show substantial technical diversity, but also limited evidence that robot deployment produces measurable improvements in routine clinical work \cite{marmaglio2023,cruz2024,babalola2024,kim2026}. This distinction is important for logistics applications. A robot may complete a transport mission successfully without reducing staff workload if people must still perform loading, receiving, access control, or recovery tasks.

Research on hospital material transport has examined both system design and operational performance. \"Ozkil and colleagues analyzed existing hospital transport processes before defining requirements for a mobile service robot \cite{ozkil2009}. Fragapane and colleagues evaluated autonomous mobile robots for sterile instrument logistics and showed that the suitability of a transport system depends on its fit with the hospital layout and material flow \cite{fragapane2023}. Bhosekar and colleagues linked inventory decisions with hospital material handling through discrete event simulation \cite{bhosekar2023}. Ge and colleagues used the same methodological approach to examine automated guided vehicle systems in operating theaters \cite{ge2025}. These studies establish simulation as an appropriate method for evaluating hospital transport systems, while also showing that demand structure, layout, routing, and operational rules must be represented explicitly.

Existing models often treat vehicle movement as the principal source of capacity. That abstraction is suitable for evaluating routing and fleet size, but it can overstate autonomy when mission completion depends on staff or shared facilities. Elevators, charging stations, cleaning facilities, and handoff activities can create queues that are not visible in a travel centered model. The present study therefore represents robot movement as one component of a coupled service process rather than as an independent replacement for human transport.

\subsection{Task allocation and fleet control}

Robot dispatch belongs to the broader class of task allocation problems. Gerkey and Matari\'c formalized this class in terms of robot capabilities, task characteristics, temporal structure, and allocation objectives \cite{gerkey2004}, while learning-based HRI research has also explored graph-based representations of human-generated information, including natural-language interactions \cite{swietlicka2024graph}. Aziz and colleagues reviewed allocation methods for robot teams and emphasized the relation between decision structure, uncertainty, and performance criteria \cite{aziz2022}. These frameworks indicate that a dispatch policy should identify its observable state, admissible actions, and objective rather than describe assignments only through simulation scenarios.

Several studies address allocation and scheduling in settings that are directly relevant to hospital logistics. Das and colleagues developed distributed auction and consensus mechanisms for heterogeneous robots in healthcare facilities \cite{das2015}. Cheng and colleagues formulated stochastic scheduling of hospital mobile robots with time windows and operating costs \cite{cheng2023}. Verma and colleagues considered dynamic allocation and task transfer in heterogeneous robot teams \cite{verma2024}. Guzm\'an Ortiz and colleagues proposed a fleet management architecture for healthcare environments, while L\'opez and colleagues linked simulation with control software for automated transport systems \cite{guzman2021,lopez2022}. Meskar and Ahmadi-Javid incorporated explicit energy constraints into a location and routing model for autonomous delivery \cite{meskar2024}.

The methods in this literature range from global optimization and distributed negotiation to online control rules. Optimization models can coordinate assignments over a planning horizon, but they require forecasts and a tractable representation of future system behavior. Distributed methods are appropriate when individual robots negotiate tasks, although they do not by themselves represent hospital staff or building resources. Transparent online rules require fewer assumptions and support direct interpretation, but their state variables and constraints must be stated precisely. The adaptive resource aware dispatch controller examined here belongs to the last category. It makes one resource class decision at each request arrival and does not claim horizon wide optimality.

\subsection{Human work and deployment evidence}

Evidence about staff capacity is less mature than research on routing and scheduling. Babalola and colleagues found that relatively few collaborative robot studies were designed around nursing workload \cite{babalola2024}. Sommer and colleagues documented transport activities performed by nurses, but the observation of transport work does not establish that an automated trip creates a continuous interval that can be reassigned to clinical care \cite{sommer2024}. Kim and colleagues similarly emphasized the importance of clinical context and workflow alignment when evaluating hospital robots \cite{kim2026}. These findings suggest that staff outcomes should be based on scheduled work and the temporal structure of released capacity, not on robot trip counts alone.

Deployment studies also show that technical performance is shaped by local workflow. A modular logistics robot has been evaluated in three European hospital pilots \cite{dei2026}. Li and colleagues reported a six month, single center comparison of manual and robot delivery for drugs and specimens \cite{li2026}. Such studies provide useful evidence of operational feasibility, but their results depend on the selected tasks, hospital layout, handoff procedures, fleet configuration, and economic assumptions. Earlier organizational research likewise showed that robot performance depends on the social and physical environment in which a system is introduced \cite{mutlu2008}. Results from one deployment therefore cannot be transferred directly to another hospital without a comparable workflow and resource structure.

\subsection{Research gap and study position}

Prior work provides strong foundations for hospital robot design, fleet scheduling, and transport simulation. Three limitations remain relevant to claims about staff capacity. First, human loading, receiving, coordination, recovery, and cleaning support are rarely scheduled on the same workers whose capacity is expected to increase. Second, robot dispatch is not always expressed as a reproducible state dependent decision method with explicit admissibility conditions. Third, shared infrastructure and travel between consecutive assignments are often separated from the dispatch decision, even though they determine when a vehicle becomes available again.

This study addresses these limitations through a formal online controller embedded in a discrete event model of coupled human, robot, and infrastructure resources. The controller maps each arriving request and the observable system state to a human or robot assignment. The execution model then schedules the support work and shared resource use created by that decision. The contribution is therefore not a new navigation method or a globally optimal task allocator. It is a reproducible decision model for testing whether an admissible robot assignment remains beneficial after its human and infrastructure requirements are represented explicitly.

%% file: robotics_context.tex
\section{Operational Context and Modeling Scope}
\label{sec:robot-context}

\subsection{Transport missions and the limits of autonomy}
Hospital delivery robots connect clinical departments with laboratories, pharmacies, storage areas, and sterile processing services. Their value depends on completing an entire transport mission, including the transfer of material at each end. A robot should be evaluated against the work that staff actually perform, including waiting and handoffs.

The term delivery robot covers different arrangements. A mobile base may carry a cabinet or move a cart, while a platform with a manipulator can also perform some physical transfer tasks. These arrangements allocate work differently between people and machines. A robot that travels independently can still require staff to load it, confirm receipt, resolve access problems, or recover a failed mission. The present study concerns material transport and represents loading and receiving through fixed durations. It does not simulate grasping, cart coupling, compartment access, or the availability of staff at a handoff.

\begin{figure*}[htb]
\centering
\begin{subfigure}[t]{0.48\textwidth}
    \centering
    \includegraphics[width=\linewidth]{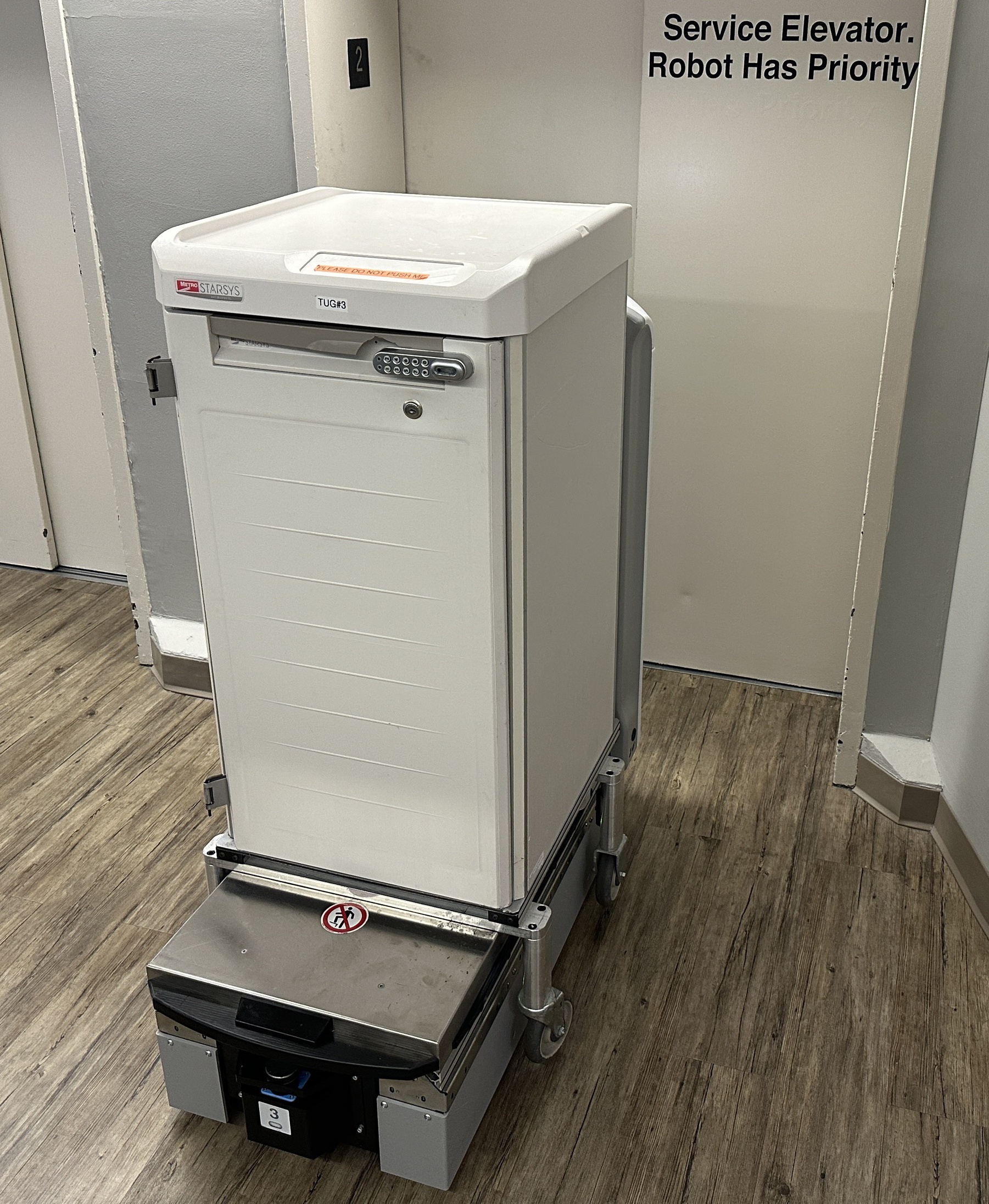}
    \caption{A hospital delivery robot waiting at service elevators marked ``Robot Has Priority'' at Danbury Hospital, Connecticut. Photograph by Mikael H\"aggstr\"om, 2024, released under CC0 1.0 \cite{danburyphoto2024}.}
    \label{fig:danbury-robot}
\end{subfigure}\hfill
\begin{subfigure}[t]{0.48\textwidth}
    \centering
    \includegraphics[width=\linewidth]{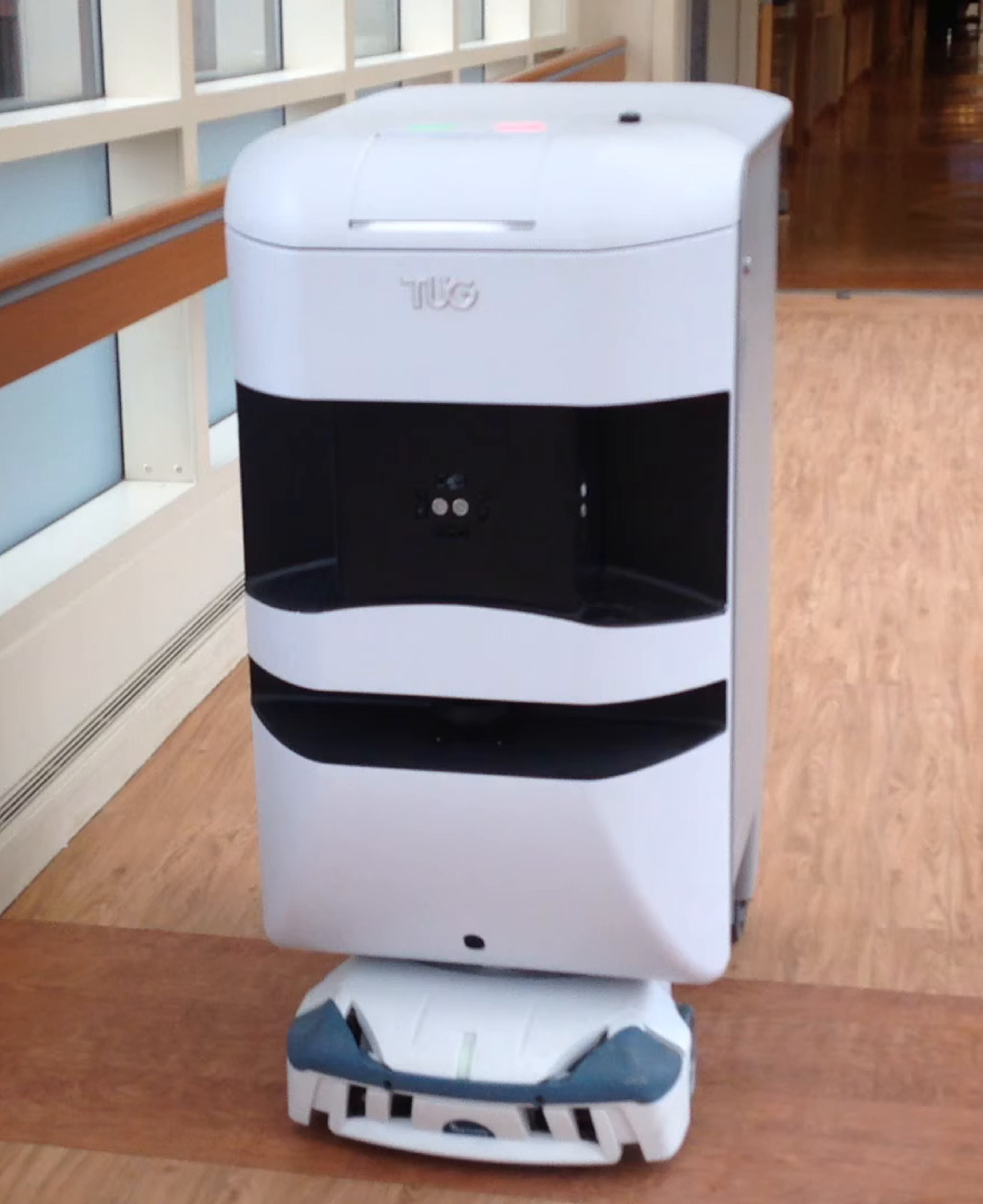}
    \caption{A TUG delivery robot operating at the University of Pittsburgh Medical Center. Still frame at 0.003 seconds from a video by Z22, 2016, CC BY-SA 4.0 \cite{tugvideo2016}. Frame extraction and cropping by the authors; the adapted image remains under CC BY-SA 4.0.}
    \label{fig:tug-robot}
\end{subfigure}
\caption{Illustrative hospital delivery robot installations. The depicted platforms were not used to calibrate or validate the simulation.}
\label{fig:hospital-robot-examples}
\end{figure*}

Figure~\ref{fig:hospital-robot-examples} shows two examples of enclosed mobile delivery robots in hospital corridors. The Danbury Hospital installation also illustrates a building level operating rule, namely priority access to a service elevator. The TUG example shows the physical scale and enclosed payload form of a robot moving through the University of Pittsburgh Medical Center. These images provide operational context only. Neither platform supplied parameters or validation data for the present simulation, and the photographs do not imply that the modeled fleet reproduces either product.

For this analysis, the relevant distinction between automated guided vehicles and autonomous mobile robots is what their control functions require from the model. A guided transport system may use a prescribed route network; a mobile robot may alter its local path around obstacles. Both still need a task assignment and access to shared facilities. Our shortest path network and stochastic travel times represent aggregate mission duration. They cannot compare navigation architectures, collision avoidance, or the quality of localization. The results apply to assumed transport service characteristics, not to a particular robot product.

\subsection{Fleet decisions and hospital infrastructure}
Robot performance has several interacting levels. Navigation determines whether a vehicle can move through a local environment. Fleet control determines which vehicle receives a request and when it becomes available again. Hospital workflow determines whether a delivery can start, where it can be received, and which staff must participate. The present model evaluates dispatch and resource availability while representing vehicle motion statistically.

Shared facilities make nominal fleet size an incomplete description of capacity. A robot waiting for an elevator is unavailable for another request even if its driving time is short. Charging and cleaning can similarly remove several vehicles from service at once. Adding robots can change the location of a queue without resolving its cause. The experiments in this article examine these mechanisms directly. They also expose what remains outside the model, including passenger elevator demand and travel from a completed delivery to the next pickup.

The controller maps each arriving task and observable hospital state to a human or robot decision. The execution model then exposes the human and infrastructure work created by that decision. This separation prevents a robot assignment from being treated as equivalent to a completed autonomous service.

\subsection{From a completed delivery to usable clinical capacity}
A completed robot delivery, released transporter time, and additional clinical care are different outcomes. The first concerns service performance. The second requires that work is removed from a person's schedule after accounting for robot support. The third also requires an appropriate role, a usable interval, and a clinical task that can be performed during that interval. Simulated transporter idle time cannot be interpreted directly as nursing time.

The capacity question concerns both robot operation and the organization of clinical work. Usable clinical capacity would require evidence that staff time remains available after handoffs and interruptions, that the person has the relevant skills, and that the receiving clinical workflow can use the time when it occurs. For a dedicated transporter, an indirect benefit might arise if reliable deliveries prevent clinical staff from leaving a ward. That pathway needs measurement of ward activity and cannot be inferred from transporter schedules alone.

The present simulation measures timeliness and gross transporter idle intervals. It records robot support burden separately. It examines part of the capacity question, while leaving the conversion to clinical work unresolved. The 10 minute threshold is an analytical choice for comparing interval structure, not an observed minimum duration for a clinical activity. This distinction guides the interpretation of all subsequent experiments.

%% file: system_architecture.tex
\begin{figure*}[htb]
\centering
\resizebox{0.99\linewidth}{!}{%
\begin{tikzpicture}[
    font=\footnotesize,
    block/.style={draw, rounded corners=2pt, align=center, minimum height=11mm,
        text width=29mm, fill=black!3},
    control/.style={block, fill=blue!9, line width=0.8pt, text width=32mm},
    resource/.style={block, fill=black!6, text width=34mm},
    flow/.style={-{Latex[length=2.2mm]}, line width=0.7pt},
    feedback/.style={-{Latex[length=2.0mm]}, dashed, line width=0.6pt}
]
\node[block] (data) {Empirical request pool\\origin, destination, time};
\node[block, right=9mm of data] (task) {Task generator\\priority, deadline, handling, route};
\node[control, right=9mm of task] (controller) {ARAD controller\\$y_i=0$: human\\$y_i=1$: robot};
\node[block, above right=8mm and 12mm of controller] (humanq) {Human priority queue\\deliveries and support};
\node[block, right=10mm of humanq] (human) {Transporter execution\\individual shifts};
\node[block, below right=8mm and 12mm of controller] (fleet) {Robot fleet manager\\idle units and batteries};
\node[block, right=10mm of fleet] (mission) {Mission execution\\load, move, receive};
\node[resource, below=10mm of mission] (resources) {Shared infrastructure\\elevators, chargers, cleaning};
\node[block, below=10mm of controller, text width=42mm] (observer) {State observer $x(t)$\\human, fleet, queue, battery, and infrastructure state};
\node[block, right=11mm of human, text width=31mm] (metrics) {Outcome recorder\\deadlines, idle intervals, support, energy};

\draw[flow] (data) -- (task);
\draw[flow] (task) -- (controller);
\draw[flow] (controller) -- node[above left]{reject robot} (humanq);
\draw[flow] (controller) -- node[below left]{admit robot} (fleet);
\draw[flow] (humanq) -- (human);
\draw[flow] (fleet) -- (mission);
\draw[flow] (human) -- (metrics);
\draw[flow] (mission) -- (metrics);
\draw[flow, bend left=18] (mission) to node[right, align=left]{support\\request} (humanq);
\draw[flow, <->] (mission) -- (resources);
\draw[feedback] (observer) -- (controller);
\draw[feedback] (humanq) |- (observer);
\draw[feedback] (fleet) |- (observer);
\draw[feedback] (resources) -- (observer);
\end{tikzpicture}%
}
\caption{System architecture and decision loop. ARAD observes queue, staff, fleet, and infrastructure state when a request arrives. The controller selects a resource class. Robot execution can generate new human support work and contention for shared infrastructure, which changes the state observed at later decision epochs.}
\label{fig:system-architecture}
\end{figure*}

%% file: dispatch_formalization.tex
\subsection{Formal controller specification}
\label{sec:formal-controller}

\subsubsection{Inputs and observable state}
Request $i$ arrives at decision time $a_i$ with the input tuple
\begin{equation}
\xi_i=(o_i,d_i,p_i,\delta_i,t_i^{L},t_i^{R},D_i,n_i^{E},n_i^{F}),
\end{equation}
where $o_i$ and $d_i$ are origin and destination, $p_i$ is priority, $\delta_i$ is the deadline, $t_i^{L}$ and $t_i^{R}$ are loading and receiving times, $D_i$ is route distance, $n_i^{E}$ is the number of elevator rides, and $n_i^{F}$ is the number of floors crossed. Priority 1 denotes urgent work. Remaining slack at time $t$ is $s_i(t)=\delta_i-t$.

The event driven system state is

%\begin{equation}
%x(t)=\bigl(H_a,H_b,Q_H,A,F,\mathbf{b},\mathbf{l},N_E,Q_E,C_E,
%N_C,Q_C,\\ C_C,N_D,Q_D,C_D\bigr),
%\end{equation}

\begin{equation}
\begin{split}
x(t) = \bigl( & H_a, H_b, Q_H, A, F, \mathbf{b}, \mathbf{l}, N_E, Q_E, C_E, \\
              & N_C, Q_C, C_C, N_D, Q_D, C_D \bigr),
\end{split}
\end{equation}

where $H_a$ and $H_b$ are active and busy transporters, $Q_H$ is the human queue length, $A$ is the number of idle robots, and $F$ is fleet size. Vectors $\mathbf{b}$ and $\mathbf{l}$ contain robot battery states and locations. For elevators ($E$), chargers ($C$), and decontamination stations ($D$), $N$, $Q$, and $C$ denote occupied servers, waiting requests, and capacity. The controller uses aggregate availability and pressure rather than selecting a particular robot. The first idle robot in the fleet store executes an admitted mission. Battery state is checked by the execution layer after the unit is acquired. Robot location enters the duration prediction only in the repositioning stress experiment; repositioning is disabled in the primary analysis.

The number of idle humans and three normalized pressure measures are
\begin{align}
H_0(t)&=\max\{0,H_a(t)-H_b(t)\},\\
P_H(t)&=Q_H(t)/\max\{1,H_a(t)\},\\
P_E(t)&=\{N_E(t)+Q_E(t)\}/C_E,\\
P_S(t)&=\max\!\left\{\frac{N_C(t)+Q_C(t)}{C_C},
                         \frac{N_D(t)+Q_D(t)}{C_D}\right\}.
\end{align}
These variables are sampled when the request arrives. A request sent to the human queue is not reconsidered later.

Let $t_k$ be the $k$th request arrival and let $i_k$ identify the arriving request. The controller is the deterministic state feedback map

%\begin{equation}
%y_{i_k}=\pi_{\mathrm{ARAD}}\!\left(\xi_{i_k},x(t_k^-)\right),
%\qquad
%x(t_{k+1}^-)=\Phi\!\left(x(t_k^-),\xi_{i_k},y_{i_k},\omega_k\right),
%\end{equation}

\begin{equation}
\begin{split}
y_{i_k} &= \pi_{\mathrm{ARAD}}\!\left(\xi_{i_k},x(t_k^-)\right), \\
x(t_{k+1}^-) &= \Phi\!\left(x(t_k^-),\xi_{i_k},y_{i_k},\omega_k\right),
\end{split}
\end{equation}

where $\Phi$ is the event driven simulation transition and $\omega_k$ contains the stochastic service, congestion, and failure draws realized before the next request arrival. The decision map is reproducible for a given request and observed state. Randomness enters through the subsequent system transition, not through the assignment rule.

\subsubsection{Duration prediction and mission state}
The deterministic origin to destination estimate is
\begin{equation}
\widehat T_i=t_i^{L}+\frac{D_i}{v_R}+t_E n_i^{E}+t_F n_i^{F}+t_i^{R},
\label{eq:robot-duration-estimate}
\end{equation}
with $v_R=54$ m/min, $t_E=2.2$ min, and $t_F=0.25$ min. Let $T_i^{\mathrm{rep}}(\mathbf l)$ be deterministic travel from the location of the first idle robot to $o_i$, and let $z_{\mathrm{rep}}$ indicate whether repositioning is enabled. The controller prediction is
\begin{equation}
\widehat T_i^{C}(t)=\widehat T_i+z_{\mathrm{rep}}T_i^{\mathrm{rep}}(\mathbf l(t)).
\end{equation}
Elevator pressure produces the state adjusted estimate
\begin{equation}
\widetilde T_i(t)=\widehat T_i^{C}(t)\{1+\alpha P_E(t)\}, \qquad \alpha=0.15.
\label{eq:state-adjusted-duration}
\end{equation}
This inflation is a controller parameter, not a fitted elevator delay model.

The primary analysis sets $z_{\mathrm{rep}}=0$. At execution, congestion is sampled and produces an operational mission duration $T_i^{\mathrm{exec}}$. Required energy is $e_i=\kappa T_i^{\mathrm{exec}}$, with $\kappa=0.025$ kWh/min. If the selected robot has less than $e_i$, it charges before loading. A mission is infeasible if $e_i$ exceeds the 4 kWh full battery capacity. No such infeasible mission occurred in the primary experiments.

Each robot follows the discrete mission mode
%\begin{equation}
%m_r\in\{\text{idle},\text{precheck},\text{charge},\text{loading},
%\text{moving},\text{receiving},\text{cleaning},\text{release}\}.
%\end{equation}

\begin{equation}
\begin{split}
m_r \in \{ & \text{idle}, \text{precheck}, \text{charge}, \text{loading}, \text{moving}, \text{receiving}, \\
& \text{cleaning}, \text{release} \}.
\end{split}
\end{equation}

Admission changes an idle robot to precheck. If its current energy is below the planned mission requirement, it charges before loading. Full battery energy must cover the mission. Loading, coordination, and any recovery intervention occupy a transporter through the human queue. Movement can occupy the shared elevator bank. Receiving creates another human support task. Specimen and instrument missions then enter cleaning. A robot with no more than 25 percent charge enters charging; otherwise it returns to the idle store. These event transitions define when $A$, $\mathbf b$, $\mathbf l$, and the shared resource states change.

\subsubsection{Admissibility and assignment decision}
Let the configured reserve be $r_0=1$. The fleet adjusted reserve is
\begin{equation}
R(F)=\min\{r_0,\max(0,F-1)\},
\end{equation}
and let $\rho_H=0.25$ and $\rho_S=1.0$ be the human and support pressure thresholds. The common feasibility gate is
\begin{equation}
g_i^0(t)=\mathbf{1}\{A(t)>0\}\mathbf{1}\{H_0(t)=0\}
         \mathbf{1}\{\widetilde T_i(t)\leq s_i(t)\}.
\end{equation}
For urgent and nonurgent work, respectively,
\begin{align}
g_i^{U}(t)&=g_i^0(t),\\
g_i^{N}(t)&=g_i^0(t)\mathbf{1}\{P_H(t)\geq\rho_H\}
 \mathbf{1}\{P_S(t)\leq\rho_S\}\mathbf{1}\{A(t)>R(F)\}.
\end{align}
The binary assignment is
\begin{equation}
y_i(t)=
\begin{cases}
1, & p_i=1\ \text{and}\ g_i^{U}(t)=1,\\
1, & p_i>1\ \text{and}\ g_i^{N}(t)=1,\\
0, & \text{otherwise},
\end{cases}
\label{eq:arad-decision}
\end{equation}
where $y_i=1$ assigns the request to the robot class and $y_i=0$ places it in the human queue. Urgent work is not constrained by the reserve or pressure thresholds. The deadline feasible comparator uses $y_i=g_i^0$ for every priority.

For each arrival, ARAD is equivalently the one step constrained decision
%\begin{equation}
%\max_{y_i\in\{0,1\}} y_i
%\quad\text{subject to}\quad
%y_i\leq g_i^{U}(t)\ \text{if }p_i=1,
%\quad y_i\leq g_i^{N}(t)\ \text{if }p_i>1.
%\end{equation}

\begin{equation}
\begin{split}
\max_{y_i\in\{0,1\}} y_i
\text{subject to} \\
& y_i\leq g_i^{U}(t)\ \text{if }p_i=1, \\
& y_i\leq g_i^{N}(t)\ \text{if }p_i>1.
\end{split}
\end{equation}

Thus, its local objective is to admit a robot mission whenever all applicable conditions hold. It does not minimize a forecast cost over future arrivals and does not claim global optimality.

\begin{algorithm}[htb]
\caption{Adaptive resource aware dispatch at a request arrival}
\label{alg:arad}
\begin{algorithmic}[1]
\Require Request $\xi_i$, observed state $x(t)$, parameters $(\alpha,\rho_H,\rho_S,r_0)$
\State Compute $H_0$, $A$, $P_H$, $P_E$, $P_S$, and $s_i$
\State Compute $\widehat T_i$ from Eq.~\eqref{eq:robot-duration-estimate}
\State $\widehat T_i^{C}\gets\widehat T_i+z_{\mathrm{rep}}T_i^{\mathrm{rep}}(\mathbf l)$
\State $\widetilde T_i\gets\widehat T_i^{C}(1+\alpha P_E)$
\If{$A=0$ \textbf{or} $H_0>0$ \textbf{or} $\widetilde T_i>s_i$}
    \State \Return human queue
\EndIf
\If{$p_i=1$}
    \State \Return robot fleet
\EndIf
\If{$P_H<\rho_H$ \textbf{or} $P_S>\rho_S$ \textbf{or} $A\leq R(F)$}
    \State \Return human queue
\EndIf
\State \Return robot fleet
\end{algorithmic}
\end{algorithm}

\subsubsection{Evaluation objectives}
Controller parameters were fixed before the reported comparison and were not tuned against the outcome replications. Policies are evaluated with the objective vector
\begin{equation}
\mathbf{J}(\pi)=\left(L_{\rm all},L_{\rm urgent},-I_{\tau},S_H,E_R,W_E,N_I\right),
\end{equation}
where $L_{\rm all}$ and $L_{\rm urgent}$ are late or incomplete fractions, $I_{\tau}$ is contiguous human idle time, $S_H$ is human support time, $E_R$ is modeled robot mission energy, $W_E$ is elevator waiting, and $N_I$ is the number of human interventions. No scalar weighting is imposed, because a reduction in one quantity cannot be assumed to compensate for worse urgent service or additional human work. Primary inference concerns the first three outcomes; the remaining terms explain mechanisms.

%% file: model_details.tex
\subsection{Stochastic service assumptions}
Distance travel is multiplied by $\max(0.65,\exp(Z))$ with $Z\sim N(-\sigma^2/2,\sigma^2)$ and reference $\sigma=0.18$. Elevator and handling time are then added. Failures are independent Bernoulli mission events and recovery is exponential. These are assumed distributions, not fitted robot telemetry. All task types are robot eligible subject to policy rules; payload, specimen packaging, and compatibility with specific routes are not modeled.

\subsection{Reference parameters and experiment design}
\label{sec:technical}
Table~\ref{tab:reference-parameters} summarizes the reference inputs. Structural variants change selected infrastructure assumptions as described below.

\begin{table}[htb]
\centering\small
\caption{Reference model inputs. Values are assumptions unless identified as empirical inputs in the text.}
\label{tab:reference-parameters}
\begin{tabular}{p{0.34\linewidth}p{0.19\linewidth}p{0.37\linewidth}}
\toprule
Parameter & Reference value & Interpretation \\
\midrule
Hospital beds & 200 & Normalization denominator \\
Warmup period & 7 days & Excluded from outcome measurement \\
Measurement period & 30 days & Included in every principal run \\
Night staffing & 2 transporters & 22:00 to 06:00 \\
Day staffing & 8 transporters & 06:00 to 14:00 \\
Evening staffing & 7 transporters & 14:00 to 22:00 \\
Human speed & 72 m/min & Reference walking speed \\
Robot speed & 54 m/min & Reference autonomous speed \\
Elevator cycle & 2.2 min & Door, access, and fixed ride component \\
Elevator floor time & 0.25 min & Added for each floor crossed \\
Robot reliability & 0.98 & Probability of mission without failure \\
Failure recovery & 12 min & Mean exponential recovery duration \\
Coordination scale & 2 min & Lognormal scale; mean is 2.09 min before recovery \\
Battery energy use & 0.025 kWh/min & Planned mission duration \\
Charge threshold & 25 percent & Charging initiation level \\
Charging efficiency & 0.90 & Grid to battery efficiency \\
Idle interval threshold & 10 min & Minimum contiguous human interval \\
\bottomrule
\end{tabular}
\end{table}

\subsubsection{Experiment inventory}
The principal policy comparison includes all modeled infrastructure constraints. Earlier experiments establish staffing and explore individual mechanisms under different configurations. Their effects are not additive. Table~\ref{tab:experiment-inventory} reports execution counts, including repeated human baselines; these are not independent clinical observations. All capacity labels retained in archived figures refer to gross transporter idle intervals.

\begin{table*}[htb]
\centering\small
\caption{Experiment inventory. Execution counts include repeated human baselines.}
\label{tab:experiment-inventory}
\begin{tabular}{p{0.29\linewidth}r p{0.49\linewidth}}
\toprule
Experiment & Simulations & Principal purpose \\
\midrule
Staffing calibration & 5,040 & Select shift counts meeting service and utilization targets \\
Robot policy comparison & 8,100 & Compare human, overflow, and priority policies \\
Global sensitivity & 4,096 & Vary nine uncertain inputs simultaneously \\
Elevator contention & 2,400 & Test shared elevator capacity and service time \\
Battery and charging & 2,700 & Test energy, charger count, and charger power \\
Exact route batching & 1,400 & Test collection windows and batch capacity \\
Cleaning downtime & 1,400 & Test cleaning duration and station count \\
Layout sensitivity & 1,000 & Compare five virtual hospital structures \\
Mixed dispatch & 1,200 & Compare overflow, priority, and adaptive dispatch \\
\midrule
Total listed simulation runs & 27,336 & Excludes small verification and development runs \\
\bottomrule
\end{tabular}
\end{table*}

\subsubsection{Empirical route characteristics}

\begin{table*}[htb]
\centering
\caption{Distribution of the 1,913 empirical origin and destination routes.}\label{tab:empirical-routes}
\begin{tabular}{lrrrrr}
\toprule
Measure & Minimum & Median & Mean & 95th percentile & Maximum \\
\midrule
Distance, m & 14.4 & 298.0 & 317.0 & 675.9 & 877.7 \\
Elevator rides & 0 & 1 & 1.21 & 2 & 3 \\
Floors crossed & 0 & 2 & 2.86 & 8 & 13 \\
Human travel, min & 0.45 & 7.49 & 7.78 & 13.52 & 18.07 \\
Robot travel, min & 0.60 & 8.83 & 9.25 & 16.34 & 22.12 \\
\bottomrule
\end{tabular}
\end{table*}

Table~\ref{tab:empirical-routes} describes the empirical routes. Elevators occur in 88.2 percent of them. These distributions define
descriptive targets for travel and infrastructure variants.

Table~\ref{tab:global-ranges} gives the exploratory input ranges.

\begin{table}[htbp]
\centering
\caption{Latin hypercube input ranges.}\label{tab:global-ranges}
\begin{tabular}{lrr}
\toprule Input & Low & High \\
\midrule
Demand multiplier & 0.75 & 1.25 \\
Robot reliability & 0.950 & 0.995 \\
Coordination time, min & 0.5 & 4.0 \\
Congestion sigma & 0.10 & 0.35 \\
Elevator cycle, min & 1.2 & 6.0 \\
Human speed, m/min & 55 & 90 \\
Robot speed, m/min & 35 & 75 \\
Handling time multiplier & 0.75 & 1.25 \\
Idle interval threshold, min & 5 & 30 \\
\bottomrule
\end{tabular}
\end{table}

\subsubsection{Structural scenarios}

The elevator experiment used banks of one, two, and four servers and cycle times
of 1.2, 2.2, and 4.0 minutes. The battery experiment used 2 and 4 kWh batteries,
one or two chargers, and power of 1 or 2 kW. The cleaning scenarios used
cycles of 3, 7, and 12 minutes and one or two stations. Batch windows were 2, 5,
and 10 minutes with capacities of two or four requests.

The compact layout multiplied distance by 0.70. The horizontal layout multiplied
distance by 1.30 and removed elevator travel. The sprawling layout multiplied
distance by 1.60. The vertical layout multiplied distance by 0.90, required at
least one elevator ride for every route, and increased rides and floors.

%% file: revision_results.tex
\subsection{Staffing recalibration}

The revised human roster was selected before the robot comparison. We screened 48 schedules with five replications each, then evaluated the schedules at 128 and 136 staffed hours per day in 30 replications. No 128 hour schedule met both lateness targets. Four 136 hour schedules did. The selected schedule has two night, eight day, and seven evening transporters. Its confirmation means were 6.63 percent late or incomplete, 3.89 percent urgent late or incomplete, zero incomplete requests, and human utilization of 0.470. The original one, seven, and six schedule was not used as the primary comparator.

\subsection{Revised primary comparison}

Scheduling robot support changed the direction of the result. Table~\ref{tab:revision-primary} shows that every tested robot fleet had higher overall and urgent lateness than the paired human baseline. With two ARAD robots, overall lateness increased by 0.37 percentage points (95 percent Monte Carlo confidence interval, 0.26 to 0.49), and urgent lateness increased by 0.82 points (0.70 to 0.95). Human idle time in intervals of at least 10 minutes decreased by 11.5 minutes per 100 nominal beds per day ($-12.9$ to $-10.1$). The corresponding two robot changes for the deadline feasible policy were 0.31 points (0.22 to 0.41), 0.60 points (0.50 to 0.71), and $-10.0$ minutes ($-11.5$ to $-8.4$).

\begin{table*}[htb]
\centering\small
\caption{Revised primary results from 30 paired replications. Idle and support values are minutes per 100 nominal beds per measured day.}
\label{tab:revision-primary}
\begin{tabular}{llrrrr}
\toprule
Policy & Fleet & Late or incomplete (\%) & Urgent late (\%) & Idle $\geq10$ min & Human support \\
\midrule
Human only & 0 & 6.60 & 3.79 & 2068.3 & 0.0 \\
ARAD & 1 & 6.83 & 4.21 & 2059.7 & 29.4 \\
ARAD & 2 & 6.98 & 4.62 & 2056.8 & 56.4 \\
ARAD & 4 & 7.42 & 5.44 & 2052.7 & 100.1 \\
Deadline feasible & 1 & 6.85 & 4.20 & 2062.0 & 29.5 \\
Deadline feasible & 2 & 6.92 & 4.40 & 2058.4 & 54.3 \\
Deadline feasible & 4 & 7.47 & 5.21 & 2052.8 & 96.4 \\
\bottomrule
\end{tabular}
\end{table*}

\begin{figure*}[htb]
\centering
\includegraphics[width=0.92\linewidth]{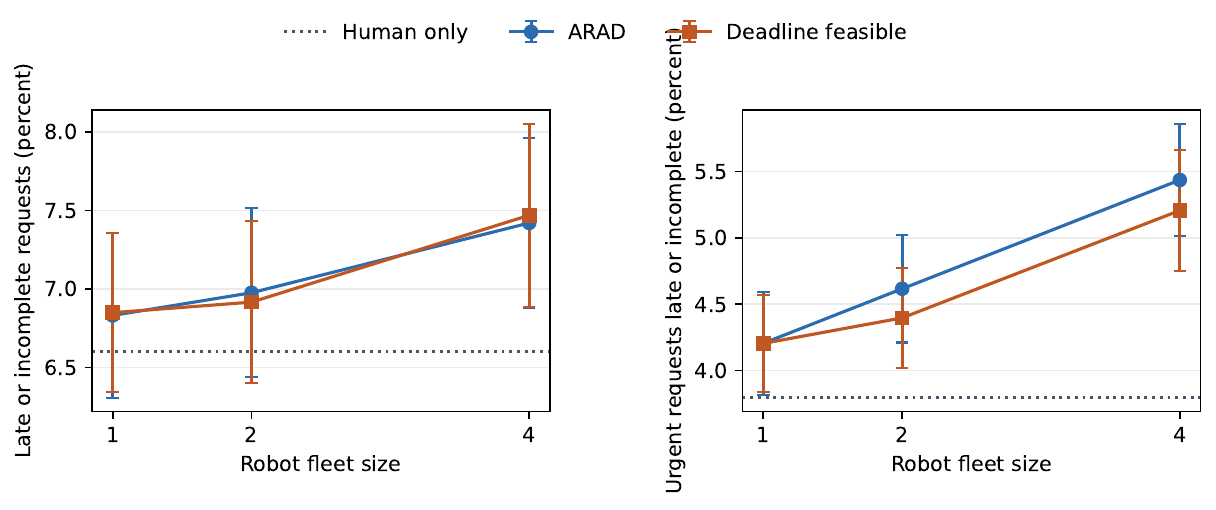}
\caption{Primary deadline outcomes. Points are means and bars are 95 percent Monte Carlo confidence intervals across 30 replications. The dotted line is the paired human only mean.}
\label{fig:revision-primary}
\end{figure*}

The larger fleets performed more robot missions, but each mission also generated scheduled loading, receiving, coordination, recovery when needed, and cleaning support. Mean support demand rose from about 29 minutes with one robot to 96--100 minutes with four robots on the normalized daily scale. Mean support waits were 1.86 minutes for two ARAD robots and 2.80 minutes for two deadline feasible robots. Battery admission checks triggered no forced charge before a mission, and no infeasible mission occurred. The result is driven by human support and queue interactions rather than by an energy failure.

\subsection{Policy ablation and stronger benchmark}

Removing individual ARAD gates produced small changes. Mean overall lateness was 7.25 percent for the reference controller, 7.12 percent without the robot reserve, 7.20 percent without the human pressure gate, and 7.17 percent without elevator inflation. Removing the support pressure gate produced exactly the reference result, showing that this gate never bound in the tested state space. The deadline feasible policy was a stronger conceptual benchmark because it omitted the reserve and nonurgent pressure gates, but it did not outperform the human roster. The differences among robot policies were smaller than their common disadvantage relative to human only service.

\subsection{Task mapping and urgency sensitivity}

The synthetic mapping materially affected absolute results and the policy contrast. With the source event mix and fixed priorities, mean lateness was 6.72 percent for humans, 6.99 percent for ARAD, and 7.02 percent for deadline feasible dispatch. Under a balanced mix of 40 percent specimen, 30 percent medication, 20 percent supply, and 10 percent instrument work, the corresponding means were 1.68, 1.97, and 1.88 percent. When class specific mixed priorities were applied, overall lateness was 7.14, 7.09, and 7.37 percent under the source mix, but urgent lateness was 0.24, 1.42, and 0.61 percent. Under the balanced mix with mixed priorities, overall lateness was 2.88, 2.90, and 3.07 percent. The robot policies did not show a stable overall advantage across these scenarios, and urgent results changed sharply when not every specimen was urgent.

\subsection{Elevator traffic, run length, and repositioning}

Passenger elevator traffic dominated the policy effect. With six passenger requests per hour, human lateness rose from 6.50 to 14.80 percent and two robot ARAD lateness rose from 6.95 to 15.69 percent. At 12 requests per hour, the values were 33.32 and 34.86 percent. These scenarios use an assumed Poisson passenger stream and are stress tests rather than a local calibration.

Warmup and horizon changes did not alter the substantive comparison. Across zero, seven, and 14 warmup days with a 30 day measurement period, human means ranged from 6.45 to 6.54 percent and ARAD means from 6.89 to 6.96 percent. Extending measurement to 60 days gave means of 6.46 and 6.86 percent. This supports the selected seven day warmup and 30 day measurement horizon for the reported conditional means.

Explicit travel from each completed delivery to the next pickup caused severe failure under the current central priority queue. Mean late or incomplete fractions were 92.18 percent for human only service, 92.68 percent for ARAD, and 92.35 percent for deadline feasible dispatch. Human empty travel averaged 1757, 1722, and 1746 minutes per 100 nominal beds per day. The stress result shows that omitting travel to pickup is consequential. It is not a credible estimate of real performance because the model has no location aware assignment, decentralized standby rule, or observed worker starting locations. A validated movement model must combine worker positions with nearest feasible assignment rather than add travel to an unchanged central queue.

%% file: verification_results.tex
\subsection{Model Verification Evidence}
\label{sec:verification-results}

A deterministic request level audit was performed before introducing additional
structural model variants. Four complete scenarios were evaluated: the human
baseline, the reference two robot overflow policy, a low demand scenario with
robot reliability of 0.95, and a high demand scenario with reliability of
0.995. Each scenario was evaluated with 12 conservation and accounting checks.
All 48 checks passed.

For every scenario, the number of requests created equaled the number completed
plus the number remaining incomplete. Every completed request had exactly one
event record and one transport mode. Human and robot mode counts agreed with the
aggregate simulation outputs. Service started after creation and completion followed service start. Deadlines were derived from creation plus the task allowance; completion could occur after the deadline. Event level duration, waiting
time, cycle time, lateness, robot failure, and human intervention totals agreed
with the aggregate outcome measures. No audited scenario contained an
incomplete request.

A reproducible 200 request event trace was saved for the reference two robot
overflow scenario. The trace contains task identity, request class, origin,
destination, priority, transport mode, creation time, deadline, service start,
completion time, service duration, lateness status, and human worker identity
when applicable.

The verification evidence supports internal accounting correctness. It does not
establish external validity for a particular hospital. The mapping from clinical
events to logistics requests and the operational interpretation of released
staff time still require review by a hospital operations expert.

%% file: supplement.tex
\appendix
\section{Component Experiments Retained for Transparency}
\label{app:component-experiments}

The primary comparison in the main text supersedes the component estimates below. These experiments document how staffing and individual robot constraints developed before support work was placed on worker schedules. They are retained for transparency and mechanism exploration, not as evidence for the primary ARAD effect. They use different configurations, policies, and replication counts. Their estimates cannot be added together or treated as validation of the final model.

\input{staffing_calibration.tex}
\input{mixed_dispatch_results.tex}
\input{robot_comparison_results.tex}
\input{sensitivity_results.tex}
\input{elevator_contention_results.tex}
\input{battery_charging_results.tex}
\input{batch_transport_results.tex}
\input{infection_control_results.tex}
\input{layout_sensitivity_results.tex}

%% file: staffing_calibration.tex
\subsection{Calibration of the Human Transport Baseline}
\label{sec:staffing-calibration}

The human transport baseline was calibrated by replaying complete observed days
from the Freiburg logistics dataset. The experiment preserved request times,
origins, destinations, network distances, and elevator journeys. Each simulation
contained a seven day warmup and a 30 day measurement period. Thirty replications
were performed for every staffing schedule using common demand seeds.

Three eight hour shifts were considered. The night shift covered 22:00 to 06:00,
the day shift covered 06:00 to 14:00, and the evening shift covered 14:00 to
22:00. The experiment evaluated 168 combinations and 5,040 replications in
total. A schedule was considered feasible when no more than 10 percent of all
requests were late or incomplete, no more than 5 percent of urgent requests
were late or incomplete, mean incomplete requests did not exceed 0.5 per
replication, and mean transporter utilization did not exceed 0.85.

Ten schedules met all four criteria. The schedule with the smallest number of
staffed hours contained one transporter on the night shift, seven on the day
shift, and six on the evening shift. This schedule provided 112 transporter
hours per day. Its mean late or incomplete fraction was 0.0626, its mean urgent
late or incomplete fraction was 0.0372, and its mean utilization was 0.505.
All observed requests were completed.

The result should be interpreted as a model reference schedule rather than an
estimate of the actual Freiburg workforce. The public dataset does not disclose
staff rosters, completion times, or task priorities. The calibration therefore
identifies the smallest modeled schedule that meets prespecified service
criteria under the observed request process and stated travel assumptions.

\begin{figure}[htb]
  \centering
  \includegraphics[width=0.88\linewidth]{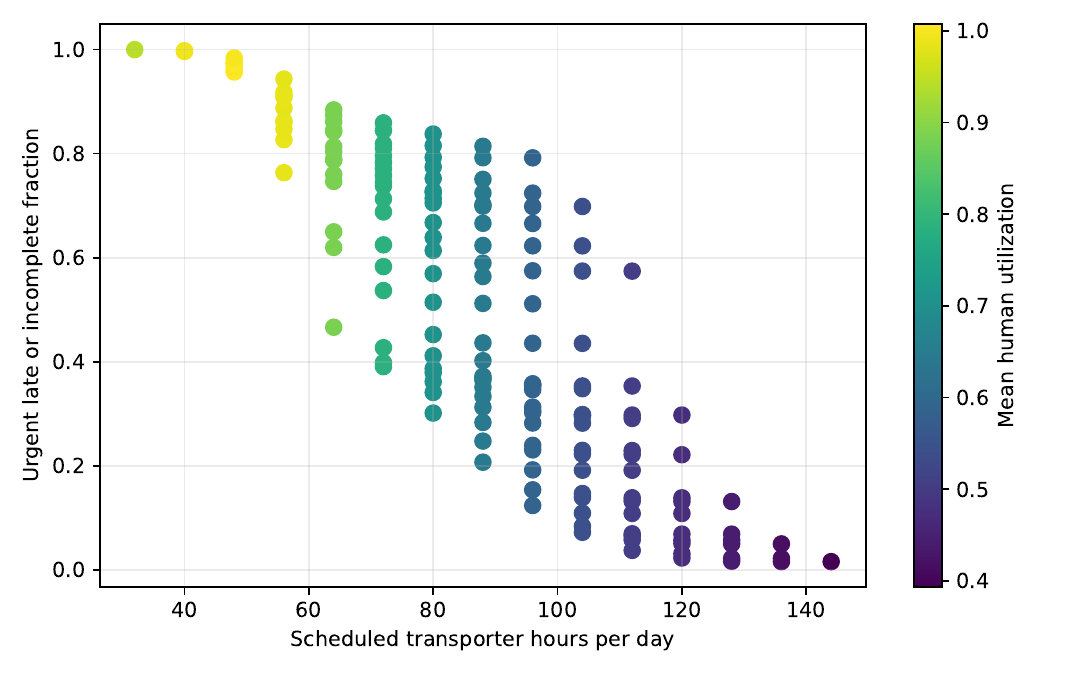}
  \caption{Urgent service failure and transporter utilization across candidate
  staffing schedules. Each point represents the mean of 30 replications.}
  \label{fig:staffing-frontier}
\end{figure}

%% file: mixed_dispatch_results.tex
\subsection{Superseded dispatch comparison without scheduled support}
\label{sec:mixed-dispatch}
In the earlier component model, before support was scheduled on workers and before staffing was recalibrated, human baseline lateness was 13.74 percent (95 percent Monte Carlo confidence interval, 13.39 to 14.08). Urgent lateness was 8.86 percent (8.56 to 9.16). These values exceed the 10 percent overall and 5 percent urgent criteria used in the earlier staffing experiment. All measurement requests completed during the clearance period. Table~\ref{tab:primary-results} reports the superseded estimates for transparency.

\input{review_primary_table.tex}

Two adaptive robots reduced overall lateness to 8.50 percent and urgent lateness to 5.32 percent. Paired reductions were 5.24 percentage points (5.10 to 5.38) and 3.54 percentage points (3.39 to 3.69). Gross idle intervals increased by 79.0 minutes per 100 nominal beds per day (77.7 to 80.4). Four adaptive robots gave overall and urgent lateness of 6.48 and 4.41 percent. These favorable estimates disappear in the revised primary model and must not be used for deployment decisions.

With two robots, adaptive dispatch reduced overall lateness by a further 0.35 percentage points relative to overflow (0.27 to 0.42), and urgent lateness by 0.55 percentage points (0.49 to 0.62). It produced 76.3 fewer additional gross idle minutes (75.0 to 77.6). Relative to priority, its additional overall lateness reduction was 0.54 percentage points (0.47 to 0.61). These direct paired comparisons support a modest timeliness advantage within the model. They do not isolate which adaptive gate produced that advantage.

Modeled mission energy over the 30 day request cohort was 255.0 kWh for two adaptive robots, 359.3 kWh for overflow, and 614.3 kWh for priority. Lower energy use partly reflects fewer robot assignments; it is not energy efficiency per delivery or total electricity consumption.

\begin{figure}[htb]
\centering
\includegraphics[width=\linewidth]{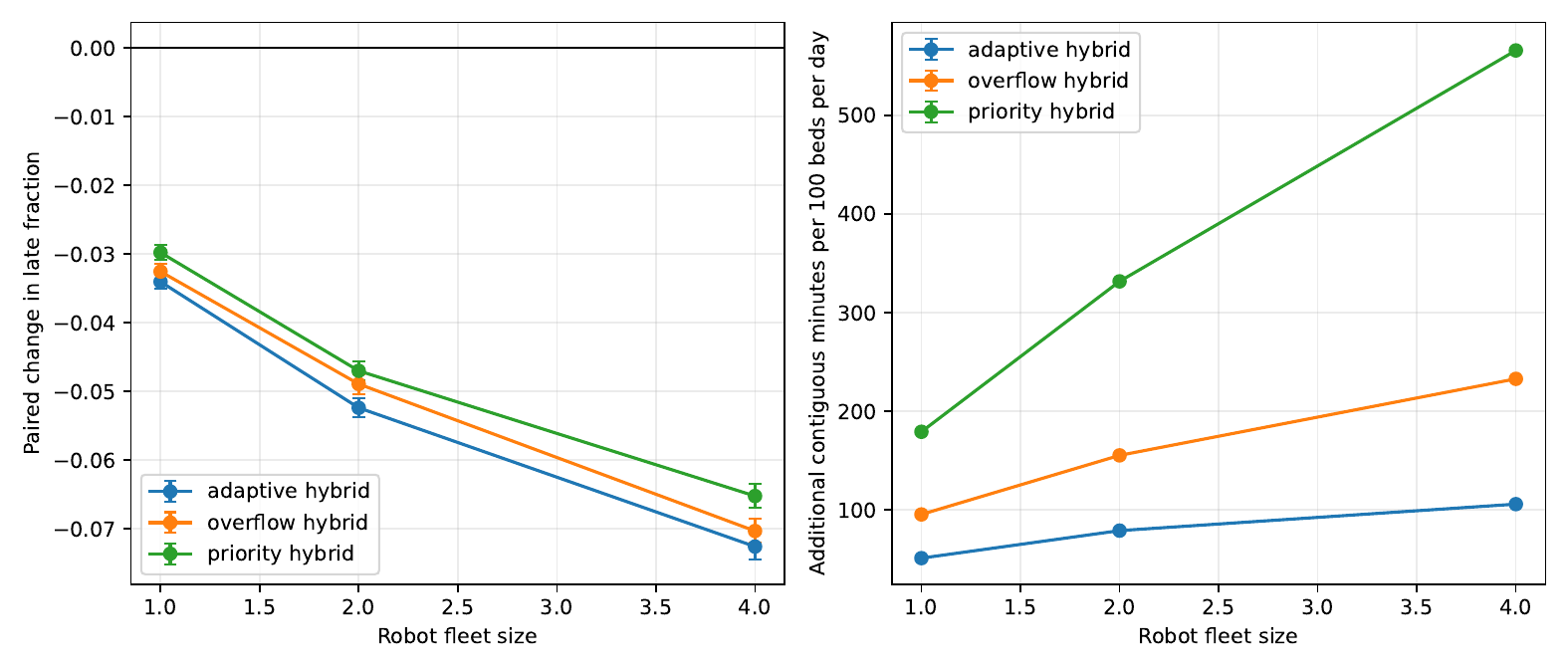}
\caption{Superseded paired changes before support scheduling and staffing recalibration. Error bars show pointwise 95 percent Monte Carlo confidence intervals. Contiguous minutes exclude robot support work.}
\label{fig:mixed-dispatch-effects}
\end{figure}

%% file: review_primary_table.tex
\begin{table*}[htb]
\centering\small
\caption{Superseded comparison before support scheduling and staffing recalibration. Absolute lateness is a percentage; changes are percentage points or gross idle minutes per 100 nominal beds per day. Brackets show pointwise 95 percent Monte Carlo confidence intervals from 100 paired replications. Gross idle intervals exclude robot support work.}
\label{tab:primary-results}
\begin{tabular}{lrrr}
\toprule
Policy, robots & Late (\%) & Late change (pp) & Gross idle change (min) \\
\midrule
Adaptive, 1 & 10.33 & -3.40 [-3.51, -3.30] & 51.1 [50.2, 52.1] \\
Adaptive, 2 & 8.50 & -5.24 [-5.38, -5.10] & 79.0 [77.7, 80.4] \\
Adaptive, 4 & 6.48 & -7.26 [-7.44, -7.08] & 105.9 [104.1, 107.7] \\
Overflow, 1 & 10.48 & -3.26 [-3.37, -3.14] & 95.4 [94.5, 96.4] \\
Overflow, 2 & 8.84 & -4.89 [-5.04, -4.74] & 155.3 [153.9, 156.8] \\
Overflow, 4 & 6.71 & -7.03 [-7.21, -6.86] & 232.9 [230.8, 235.0] \\
Priority, 1 & 10.76 & -2.98 [-3.09, -2.87] & 179.3 [178.2, 180.3] \\
Priority, 2 & 9.04 & -4.70 [-4.84, -4.56] & 331.8 [330.1, 333.4] \\
Priority, 4 & 7.21 & -6.52 [-6.69, -6.35] & 565.9 [563.4, 568.4] \\
\bottomrule
\end{tabular}
\end{table*}

%% file: robot_comparison_results.tex
\subsection{Robot Policy Comparison}
\label{sec:robot-comparison}

The calibrated human schedule was compared with priority aware hybrid dispatch
and human first overflow dispatch. The overflow policy assigned a request to a
robot only when no human transporter was idle or the human queue exceeded its
threshold, and only when a robot was immediately available. The priority aware
policy assigned a larger share of eligible requests to robots according to task
priority, deadline slack, and the estimated robot queue.

This initial model omitted explicit elevator queues, batteries, and cleaning. Its priority policy could queue work for busy robots, unlike the available robot rule in the primary comparison. It does not isolate infrastructure effects from policy implementation differences.

The experiment considered fleet sizes of one, two, four, and six robots, demand
multipliers of 0.75, 1.00, and 1.25, and robot reliabilities of 0.95, 0.98, and
0.995. One hundred replications were performed per scenario. Common demand seeds
were used for paired policy comparisons. In total, 8,100 replications were
completed. No requests remained incomplete in any reported scenario.

At reference demand and robot reliability of 0.98, the human baseline had a mean
late fraction of 0.0622 and an urgent late fraction of 0.0360. One overflow
robot reduced these outcomes to 0.0234 and 0.0139, respectively. Two overflow
robots reduced them further to 0.0117 and 0.0081. The paired reduction in the
overall late fraction was 0.0388 with one robot and 0.0505 with two robots. The
corresponding 95 percent confidence interval half widths were 0.0011 and 0.0013.

Priority aware dispatch released more nominal human transport time but performed
worse on overall timeliness with small fleets. With one robot, its late fraction
was 0.1530. With two robots it was 0.0712, which remained slightly above the
human baseline. Fleets of four and six priority aware robots reduced the late
fraction to 0.0287 and 0.0121, respectively.

The results reveal a tradeoff between service protection and the amount of work
transferred to robots. Human availability was reconstructed separately for each
scheduled transporter. An interval counted toward the idle measure only when the transporter
remained idle for at least 10 consecutive minutes. At reference demand and
reliability, one overflow robot added 144.5 gross contiguous idle minutes per 100
beds per day relative to the paired human baseline, with a 95 percent confidence
interval half width of 1.4 minutes. Two overflow robots added 233.5 minutes,
with a half width of 2.2 minutes. Four and six robots added 333.4 and 380.4
minutes, respectively. These results show diminishing marginal returns under
the overflow rule.

The priority aware policy created more contiguous staff time, but small fleets
did so at the cost of service performance. One priority aware robot added 447.9
minutes per 100 beds per day, while increasing the late fraction by 0.0908. Two
robots added 774.6 minutes, while increasing the late fraction by 0.0090. The
overflow policy provided the stronger service result for small fleets,
while the priority aware policy moved more work away from human transporters.

Robot failures generated human interventions. At reliability 0.98 and reference
demand, mean interventions ranged from 23.8 for one overflow robot to 64.0 for
six overflow robots. The corresponding range under priority aware dispatch was
62.0 to 184.8 because that policy assigned more requests to robots.

\begin{figure}[htb]
  \centering
  \includegraphics[width=0.86\linewidth]{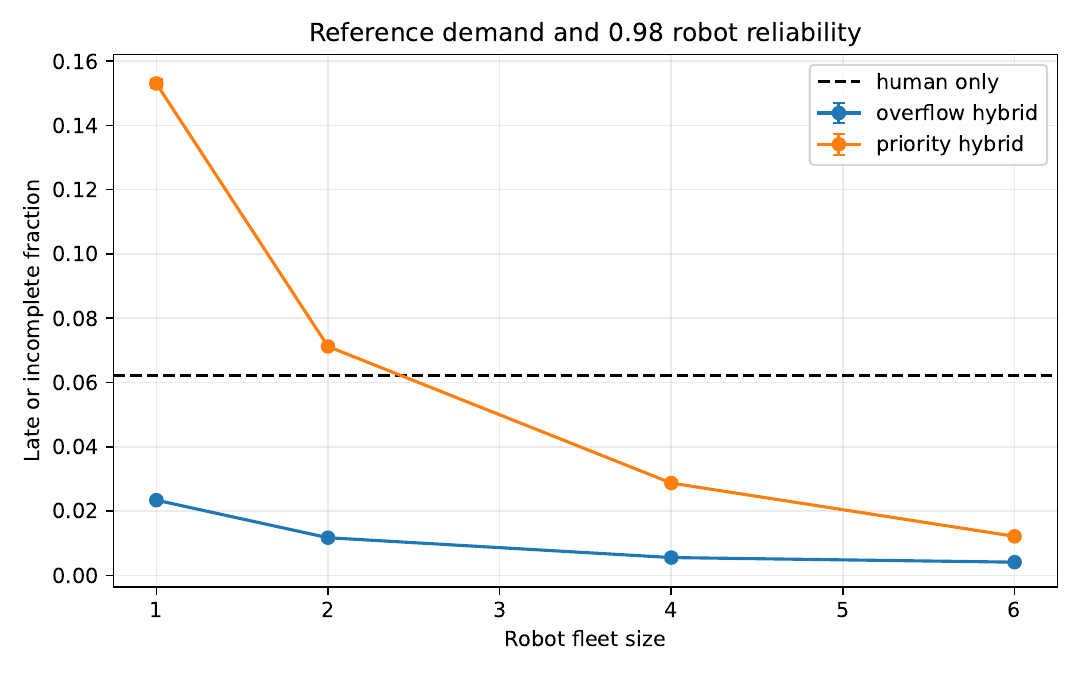}
  \caption{Mean late or incomplete fraction by dispatch policy and fleet size at
  reference demand and robot reliability of 0.98. Error bars show 95 percent
  confidence intervals.}
  \label{fig:robot-policy-comparison}
\end{figure}

\begin{figure}[htb]
  \centering
  \includegraphics[width=0.86\linewidth]{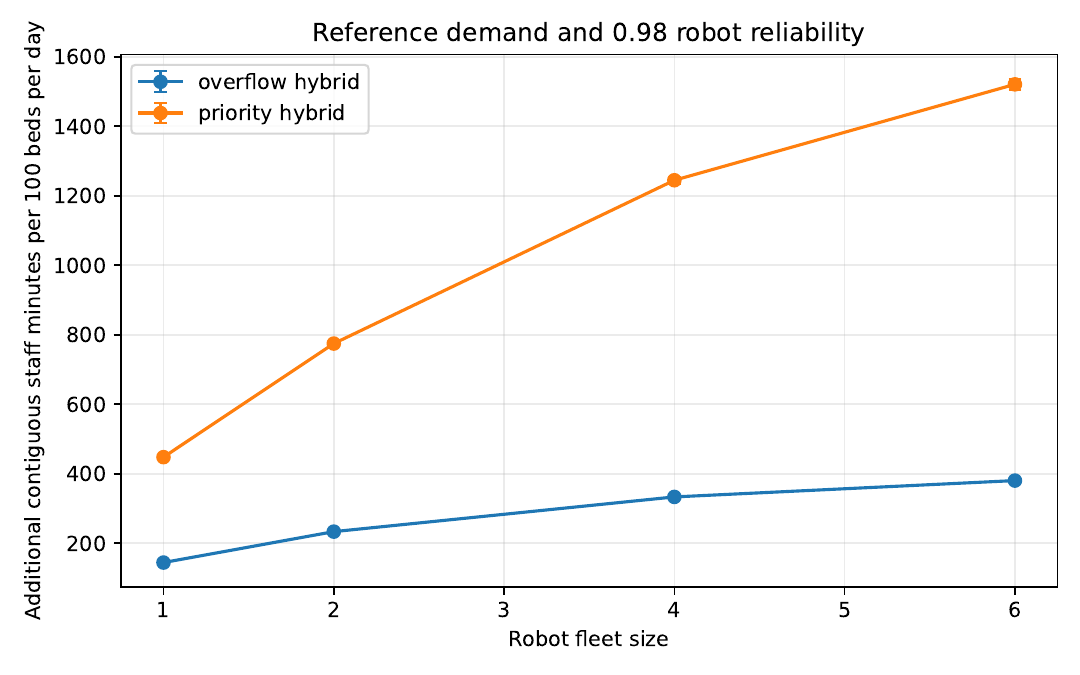}
  \caption{Paired increase in human transporter idle time occurring in
  intervals of at least 10 consecutive minutes. Values are normalized per 100
  beds per day. Error bars show 95 percent confidence intervals.}
  \label{fig:robot-contiguous-capacity}
\end{figure}

These estimates are conditional on the assumed task urgency mapping, robot
failure process, service times, calibrated human reference schedule, and the
10 minute idle interval threshold. Contiguous idle time represents gross schedule idle time that excludes robot support work, but it does not by itself establish that the time would be used
for additional work.

%% file: sensitivity_results.tex
\subsection{Global Sensitivity Analysis}
\label{sec:global-sensitivity}

The robustness of the two robot overflow policy was evaluated with a Latin
hypercube design. The design contained 512 parameter combinations and four
stochastic replications per combination. Each run was paired with a human only
baseline using common random numbers. This produced 2,048 paired observations,
or 4,096 complete hospital simulations.

Nine uncertain inputs were varied simultaneously. Demand ranged from 0.75 to
1.25 times the empirical reference level. Robot reliability ranged from 0.95 to
0.995. The design also varied coordination time, congestion variation, elevator
cycle time, human and robot travel speeds, task handling time, and the minimum
duration required for an idle interval to count as qualifying idle time. The latter
threshold ranged from 5 to 30 minutes.

Across the design, the mean reduction in the overall late fraction was 0.1680.
The median reduction was 0.1396, while the fifth and 95th percentiles of the
paired change were $-0.3736$ and $-0.0073$. Overall timeliness improved in
98.97 percent of paired runs. Urgent timeliness improved in 97.85 percent of
runs. The mean and median reductions in the urgent late fraction were 0.1490
and 0.0997, respectively.

The small set of adverse overall timeliness results contained 20 of 2,048 paired
runs. These cases had a mean demand multiplier of 0.85, mean human baseline
lateness of approximately 0.01, mean elevator cycle time of 1.78 minutes, and
mean human walking speed of 86.3 meters per minute. They describe
lightly loaded hospitals in which human transport already performs well. The
largest adverse change in overall lateness was 0.0042.

Gross contiguous idle time increased in every paired run. The mean gain was 253.8
minutes per 100 beds per day, the median was 229.8 minutes, and the fifth and
95th percentiles were 139.4 and 446.6 minutes. The most influential input for
this outcome was the minimum idle interval threshold, with a standardized rank
coefficient of $-0.552$. Elevator cycle time and demand followed, with
coefficients of 0.501 and 0.431. Longer elevator delays and higher demand create
more work that can be transferred, while a stricter definition of qualifying idle time
reduces the measured gain.

Demand and elevator cycle time were the dominant drivers of the timeliness
effect. Their standardized rank coefficients for the overall late fraction
change were $-0.625$ and $-0.596$. Human walking speed had a coefficient of
0.342. The rank regression explained 91.7 percent of variation in the overall
timeliness effect, 92.3 percent of variation in the urgent timeliness effect,
and 84.3 percent of variation in gross contiguous idle time.

\begin{figure}[htb]
  \centering
  \includegraphics[width=\linewidth]{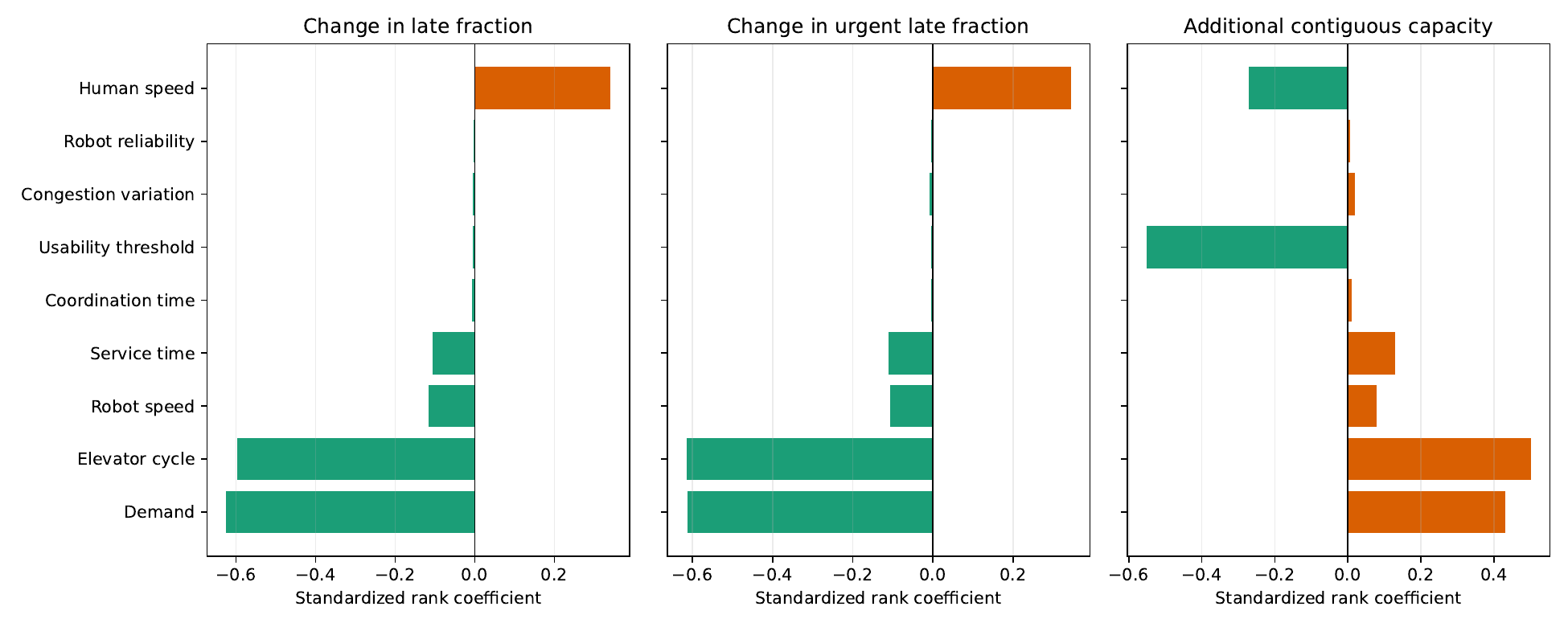}
  \caption{Standardized rank regression coefficients from the global
  sensitivity analysis. Negative timeliness coefficients indicate inputs that
  increase the reduction in lateness produced by two overflow robots.}
  \label{fig:global-sensitivity}
\end{figure}

These results describe the initial overflow model across the sampled input ranges. They do not validate the final adaptive policy. The task mix and the fraction classified as urgent were not varied. Coordination time changes a separate accounting quantity but does not occupy a worker, so this design cannot establish that idle interval gains survive human support interruptions.

%% file: elevator_contention_results.tex
\subsection{Structural Sensitivity to Elevator Contention}
\label{sec:elevator-contention}

The reference model represents elevator travel as a route specific fixed delay.
A structural sensitivity model was implemented in which human
transporters and robots request a shared elevator bank. Requests wait when all
elevators are occupied and hold an elevator for the route specific ride and
floor travel time. The experiment compared banks of one, two, and four
elevators, with elevator cycle times of 1.2, 2.2, and 4.0 minutes. Each scenario
used 100 paired replications of the human baseline and the two robot overflow
policy. The complete design contained 2,400 simulations.

With two shared elevators, the robot policy reduced the late fraction by 0.0227,
0.0723, and 0.1112 for cycle times of 1.2, 2.2, and 4.0 minutes. The corresponding
gains in gross contiguous idle time were 219.8, 211.8, and 190.7 minutes per 100
beds per day. Mean elevator waits in the human baseline ranged from 0.24 to 3.84
minutes. Under the robot policy, they ranged from 0.33 to 6.55 minutes because
robots and human transporters competed for the same resource.

With four elevators, queueing was small. The explicit contention results closely
approximated the fixed delay model at cycle times of 1.2 and 2.2 minutes. Across
the three cycle times, the two robot policy reduced the late fraction by 0.0188
to 0.1868 and added 228.1 to 280.3 gross contiguous idle minutes per 100 beds per day.

A single elevator created a bottleneck regime. At a 1.2 minute cycle time, the
robot policy changed the late fraction by 0.0013, with a 95 percent confidence
interval half width of 0.0020. This result is consistent with no timeliness
benefit. At a 2.2 minute cycle time, baseline lateness reached 0.873 and the
robot policy reduced it by 0.123, but overall service remained unacceptable.
At a 4.0 minute cycle time, baseline lateness approached 1.0 and the gain in
gross contiguous idle time fell below one minute per 100 beds per day. Robot deployment
cannot compensate for a severely undersized vertical transport system.
Incomplete requests occurred in 201 runs, all within the single elevator
scenarios at cycle times of 2.2 or 4.0 minutes. Both policies had incomplete
requests in every replication of the single elevator, 4.0 minute scenario.

\begin{figure}[htb]
  \centering
  \includegraphics[width=\linewidth]{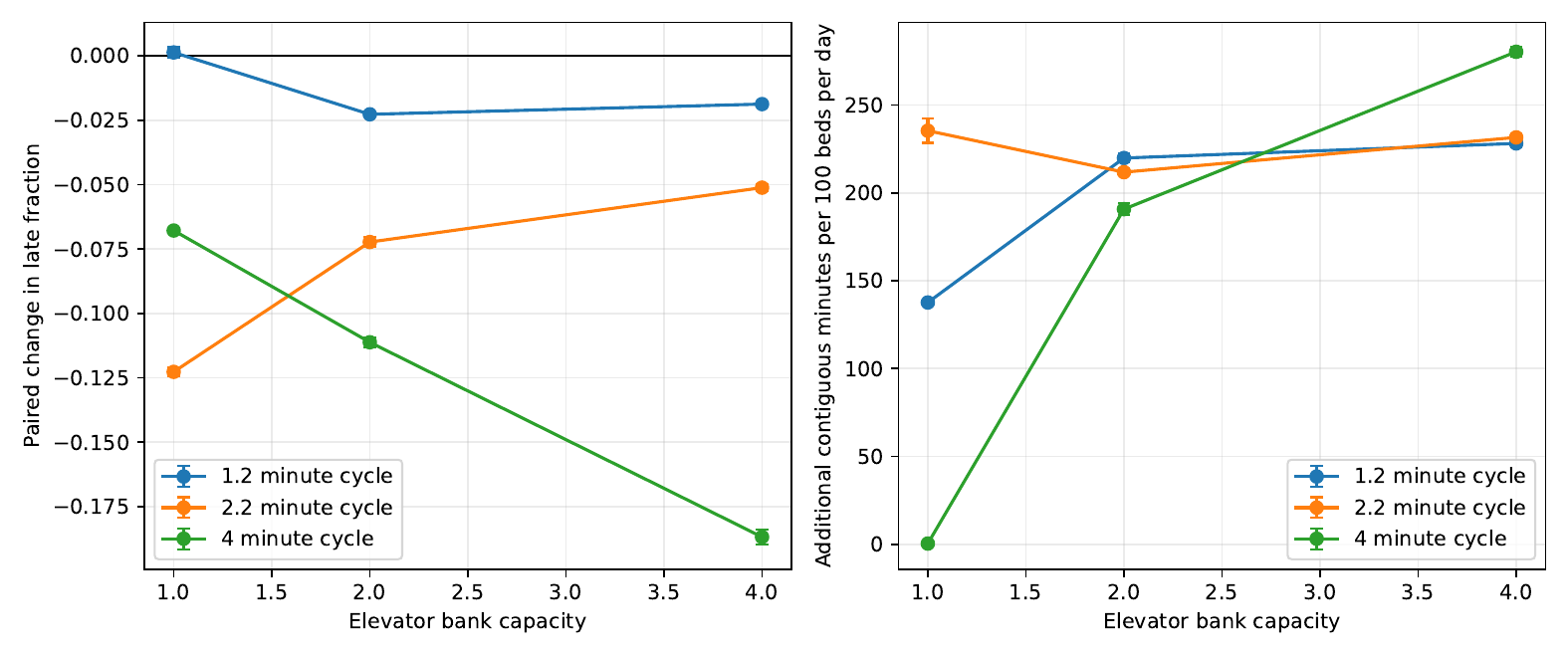}
  \caption{Paired effect of two overflow robots under explicit shared elevator
  contention. Error bars show 95 percent confidence intervals.}
  \label{fig:elevator-contention-effects}
\end{figure}

These scenarios show that elevator assumptions can change robot effects. They do not establish a universal minimum number of elevators because all route elevator service is aggregated into one shared resource request and passenger traffic is omitted.

%% file: battery_charging_results.tex
\subsection{Structural Sensitivity to Battery Charging}
\label{sec:battery-charging}

The robot model was extended with individual battery states, energy consumption,
a charge initiation threshold, and shared charging station queues. A robot began
charging when its remaining energy fell to 25 percent of capacity and returned
to service after reaching full charge. The experiment considered fleets of two,
four, and six robots, battery capacities of 2 and 4 kWh, one or two charging
stations, and charging powers of 1 and 2 kW. The assumed active energy use was
0.025 kWh per minute and charging efficiency was 0.90. Fifty paired replications
were performed per scenario, producing 2,700 simulations. No run contained an
incomplete request.

Charging reduced, but did not eliminate, the operational benefit of robots. In
the model without battery constraints, fleets of two, four, and six robots
reduced the late fraction by 0.0513, 0.0577, and 0.0593. Their corresponding
gross contiguous idle time gains were 234.5, 333.3, and 380.4 minutes per 100 beds per
day.

With two 2 kW charging stations, the strongest battery constrained scenarios
reduced the late fraction by 0.0428, 0.0531, and 0.0575 for fleets of two, four,
and six. Idle interval gains were 185.9, 281.1, and 333.8 minutes per 100 beds per day.
These configurations retained approximately 79, 84, and 88 percent of the
unconstrained idle interval benefit. The six robot configuration retained about 97
percent of its unconstrained timeliness effect.

A single 1 kW charger was inadequate for larger fleets. Mean charging waits
ranged from approximately 20 minutes for two robots with a 2 kWh battery to 487
minutes for six robots with a 4 kWh battery. Under this infrastructure, idle interval
gains fell to approximately 152 to 232 minutes per 100 beds per day. The larger
battery did not solve the problem because the full charge policy also produced
longer charging sessions.

Two charging stations removed observable queues for the two robot fleet in the
tested scenarios. For four and six robots, two 2 kW stations reduced mean waits
to approximately 3 to 21 minutes, depending on battery capacity. Charger power
and station count had a larger operational influence than battery
capacity alone.

\begin{figure}[htb]
  \centering
  \includegraphics[width=\linewidth]{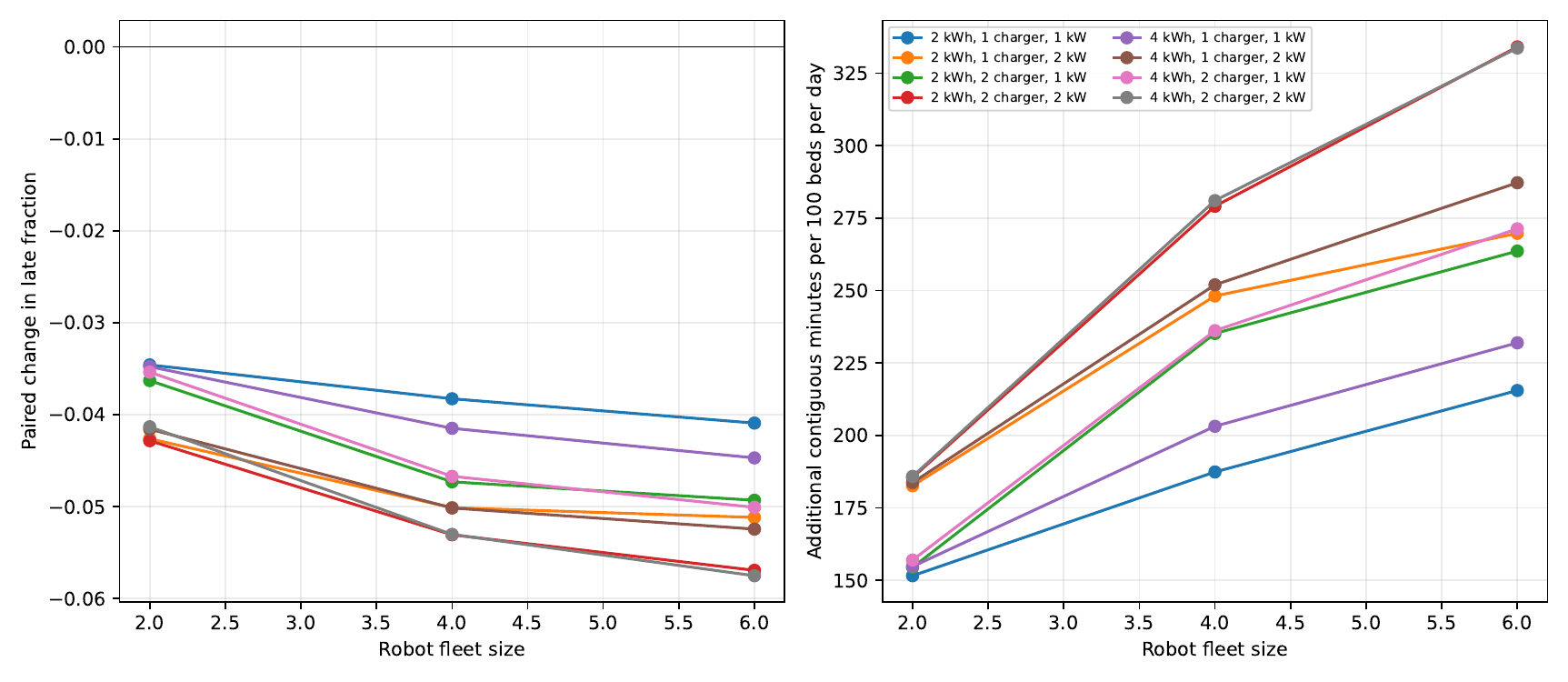}
  \caption{Paired timeliness and gross contiguous idle time effects under alternative
  battery and charging infrastructure configurations.}
  \label{fig:battery-charging-effects}
\end{figure}

Within these scenarios, charger count and power influenced robot availability. This does not establish a minimum charger specification for a hospital. The energy model excludes several operating states and does not enforce sufficient charge before mission admission.

%% file: batch_transport_results.tex
\subsection{Structural Sensitivity to Batch Transport}
\label{sec:batch-transport}

The model was extended with route compatible robot batching. Requests could
share a trip only when their origin and destination were identical. Urgent
requests were excluded from intentional batching delay. Collection windows of
2, 5, and 10 minutes and maximum batch sizes of two and four were compared with
immediate overflow dispatch. Fleets of two and four robots were evaluated with
the battery and charging model enabled. Fifty paired replications per scenario
produced 1,400 simulations. No run contained an incomplete request.

The empirical request stream provided little opportunity for exact route
consolidation. Mean realized batch size ranged from 1.005 to 1.007 requests.
Increasing maximum batch size from two to four produced no material change.
This result reflects the diversity of the 1,913 empirical origin and destination
pairs rather than an imposed robot capacity limitation.

For two robots, immediate dispatch reduced the overall late fraction by 0.0411
and the urgent late fraction by 0.0241. It added 185.9 gross contiguous idle minutes
per 100 beds per day and used 411.7 kWh over the 30 day measurement period. A
2 minute batching window reduced the overall late fraction by approximately
0.0297 and urgent lateness by 0.0113. It increased the idle interval gain to about
202.7 minutes, but robot energy use increased to approximately 442.7 kWh.

The larger idle interval estimate under batching did not result from efficient trip
consolidation. Delayed dispatch changed the overflow assignment pattern and
transferred more requests to robots, while almost every robot trip still carried
one request. The policy consumed more energy and provided less
timeliness protection.

For four robots, immediate dispatch reduced overall lateness by 0.0527 and
urgent lateness by 0.0296. A 2 minute window reduced these outcomes by 0.0449
and 0.0207. Robot energy use increased from 637.8 to approximately 664.7 kWh.
Maximum batch size again had negligible influence.

Ten minute windows performed poorly. For two and four robot fleets, urgent late
fractions increased by approximately 0.0080 and 0.0074 relative to the human
baseline. Although urgent requests were not intentionally held, delayed
nonurgent batches subsequently occupied robots and charging infrastructure.
This indirect interference harmed urgent service.

\begin{figure}[htb]
  \centering
  \includegraphics[width=\linewidth]{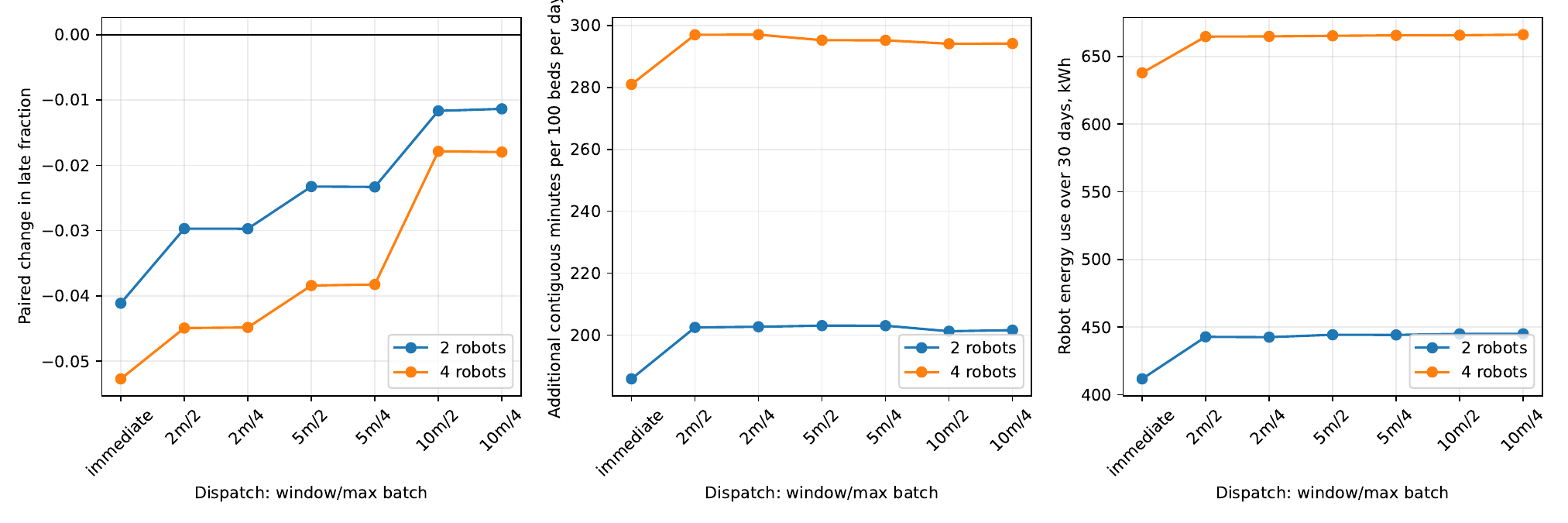}
  \caption{Effects of route compatible batching on timeliness, gross contiguous
  idle time, and robot energy consumption. Labels show batching window and
  maximum batch size.}
  \label{fig:batch-transport-effects}
\end{figure}

Immediate overflow dispatch had better timeliness and lower modeled mission energy than the tested batching strategies, while some batching scenarios produced more gross idle time. It did not dominate every outcome. The result does not rule out scheduled rounds, collection hubs, or consolidation by destination zone.

%% file: infection_control_results.tex
\subsection{Structural Sensitivity to Cleaning Downtime}
\label{sec:infection-control}

The robot model was extended with mandatory decontamination after specimen and
instrument transports. A robot remained unavailable while waiting for and
undergoing cleaning. The experiment compared cleaning durations of 3, 7, and 12
minutes, one or two decontamination stations, and fleets of two or four robots.
Battery and charging constraints remained active. Fifty paired replications per
scenario produced 1,400 simulations. No run contained an incomplete request.

Without decontamination, two robots reduced the late fraction by 0.0412 and
added 185.8 gross contiguous idle minutes per 100 beds per day. With one cleaning
station, the idle interval gain declined to 176.8, 164.2, and 152.1 minutes for
cleaning durations of 3, 7, and 12 minutes. The corresponding reductions in the
late fraction were 0.0390, 0.0377, and 0.0362. Mean station waits remained below
1.7 minutes, indicating that one station was operationally adequate for the two
robot fleet under the tested assumptions.

Without decontamination, four robots reduced the late fraction by 0.0529 and
added 280.5 gross contiguous idle minutes per 100 beds per day. With one station and a
12 minute cleaning duration, the idle interval gain fell to 217.3 minutes and the
late fraction reduction fell to 0.0451. Mean decontamination wait reached 8.22
minutes and station utilization reached 0.401.

A second cleaning station materially improved the four robot result. At a
12 minute cleaning duration, mean waiting time fell from 8.22 to 0.81 minutes
and the idle interval gain increased from 217.3 to 237.5 minutes per 100 beds per day.
The late fraction reduction increased from 0.0451 to 0.0491. At shorter cleaning
durations, the second station had a smaller effect because the single station
was not highly utilized.

\begin{figure}[htb]
  \centering
  \includegraphics[width=\linewidth]{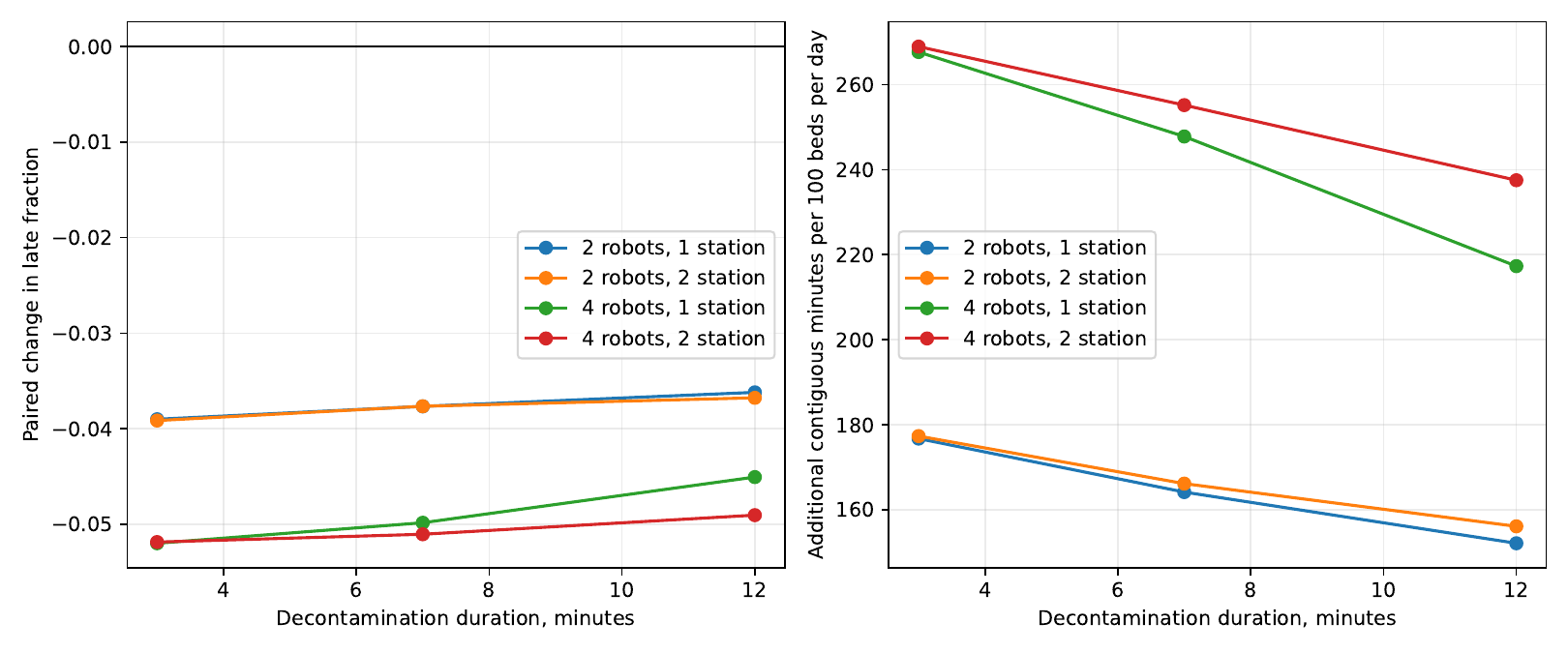}
  \caption{Effect of mandatory robot decontamination after specimen and
  instrument transports.}
  \label{fig:infection-control-effects}
\end{figure}

The modeled cleaning requirements reduced the overflow policy's idle interval gains. A second station improved the four robot scenario with a 12 minute cycle. These results concern downtime under assumed protocols, not infection risk or a validated cleaning requirement.

%% file: layout_sensitivity_results.tex
\subsection{Structural Sensitivity to Hospital Layout}
\label{sec:layout-sensitivity}

Five virtual hospital structures were evaluated while preserving the empirical
request stream. The reference layout retained the Freiburg routes. The compact
layout reduced all route distances by 30 percent. The horizontal layout
increased distance by 30 percent but removed elevator travel. The sprawling
layout increased distance by 60 percent. The vertically intensive layout
reduced horizontal distance by 10 percent but required elevator travel for every
route and increased the number of rides and floors. Two and four robot fleets
were compared with paired human baselines. Explicit elevator queues, battery
charging, and decontamination remained enabled. Fifty replications per scenario
produced 1,000 simulations.

Two robots improved overall lateness in every layout. The paired reductions were
0.0153 in the compact layout, 0.0094 in the horizontal layout, 0.0480 in the
reference layout, 0.1510 in the sprawling layout, and 0.0971 in the vertical
layout. Corresponding gross contiguous idle time gains ranged from 143.1 to 233.8
minutes per 100 beds per day.

Absolute performance is essential for interpretation. Human baseline lateness
was 0.018 in the horizontal layout and 0.057 in the compact layout. Two robots
reduced these values to 0.009 and 0.042. The smaller paired effects therefore
reflect good baseline service rather than robot failure.

In the reference structural model, baseline lateness was 0.137. Two robots
reduced it to 0.089, while four robots reduced it to 0.068. The four robot fleet
added 233.8 gross contiguous idle minutes per 100 beds per day, compared with 156.4 minutes
for two robots.

The sprawling and vertical layouts remained operationally unacceptable despite
large paired improvements. In the sprawling layout, baseline lateness was 0.521,
falling to 0.370 with two robots and 0.285 with four. In the vertical layout,
baseline lateness was 0.884, falling to 0.787 and 0.743. Mean elevator waiting
in the vertical layout increased from 5.74 minutes in the human baseline to 7.43
minutes with two robots and 9.23 minutes with four. Additional robots increased
competition for the constrained elevator bank.

Incomplete requests occurred in six unique vertical human baseline
replications. These baselines were repeated for both fleet comparisons, yielding
12 result rows with incomplete requests. No robot scenario contained an
incomplete request, but its absolute lateness remained excessive.

\begin{figure}[htb]
  \centering
  \includegraphics[width=\linewidth]{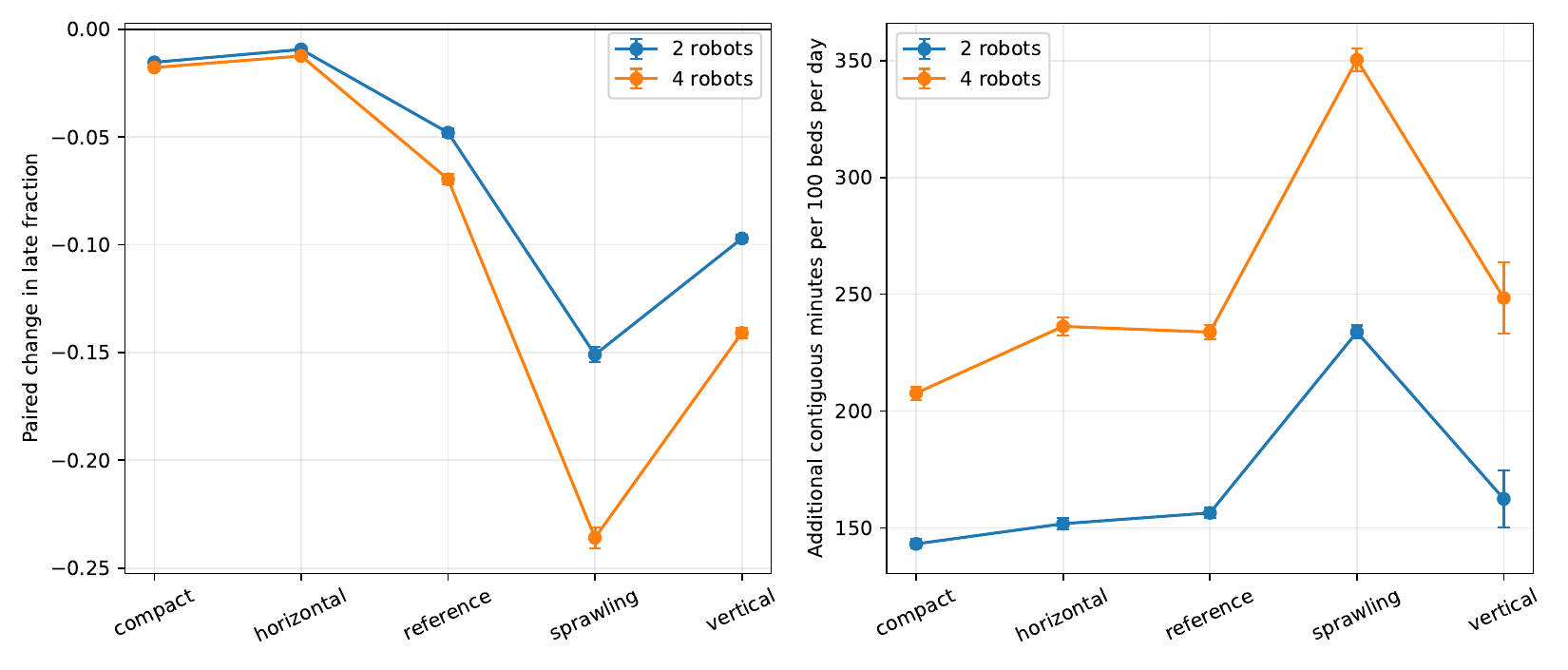}
  \caption{Paired effect of overflow robots across alternative virtual hospital
  layouts with elevator, battery, charging, and decontamination constraints.}
  \label{fig:layout-sensitivity-effects}
\end{figure}

The tested overflow fleets improved mean lateness in all five virtual layouts, but improvement did not guarantee that service targets were met. These transformations are stress scenarios derived from one network, not validation across independent hospitals. They do not identify an optimal fleet or a proven infrastructure intervention.